%% file: main.tex
\documentclass{amsart}

\usepackage[margin=1in]{geometry}
\usepackage[foot]{amsaddr}

\usepackage{url}            
\usepackage{booktabs}       
\usepackage{nicefrac}       
\usepackage{enumitem}
\usepackage[ruled]{algorithm}
\usepackage{algpseudocode}
\usepackage{pifont}
\usepackage{caption,subcaption}
\usepackage{graphicx}
\usepackage{epstopdf}
\ifpdf
\DeclareGraphicsExtensions{.eps,.pdf,.png,.jpg}
\else
\DeclareGraphicsExtensions{.eps}
\fi

\input{preamble}

\begin{document}

\title[Geodesic HMC on Stiefel manifolds for Bayesian matrix completion]{Low-rank Bayesian matrix completion via geodesic Hamiltonian Monte Carlo on Stiefel manifolds}

\author{Tiangang Cui}
\address[T. Cui]{School of Mathematics and Statistics,
  University of Sydney, NSW 2006, Australia}
\email{tiangang.cui@sydney.edu.au}

\author{Alex A. Gorodetsky}
\address[A.~A. Gorodetsky]{Department of Aerospace Engineering,
  University of Michigan,
  Ann Arbor, MI, 48109, USA}
\email{goroda@umich.edu}

\input{abstract}

\maketitle

\input{intro}

\input{sampling}

\input{examples}

\section{Acknowledgments} 
T. Cui acknowledges support from the Australian Research Council Discovery Project DP210103092.
A. Gorodetsky  acknowledges support from the Department of Energy Office of Scientific Research, ASCR under grant DE-SC0020364.
\clearpage 

\appendix

\input{appendix}

\bibliographystyle{plain}
\bibliography{references}

\end{document}

%% file: preamble.tex
\newcommand{\reals}{\mathbb{R}}
\newcommand{\mat}[1]{\mathbf{#1}}

\DeclareMathOperator*{\argmin}{arg\,min}
\DeclareMathOperator*{\argmax}{arg\,max}
\DeclareMathOperator*{\Minimize}{minimize}

\newcommand{\rev}[1]{{ #1 }}

\renewcommand{\paragraph}[1]{
    \textbf{#1 \;\;} 
}

%% file: abstract.tex
\begin{abstract}
  We present a new sampling-based approach for enabling efficient computation of low-rank Bayesian matrix completion and quantifying the associated uncertainty. Firstly, we design a new prior model based on the singular-value-decomposition (SVD) parametrization of low-rank matrices. Our prior is analogous to the seminal nuclear-norm regularization used in non-Bayesian setting and enforces orthogonality in the factor matrices by constraining them to Stiefel manifolds. Then, we design a geodesic Hamiltonian Monte Carlo (-within-Gibbs) algorithm for generating posterior samples of the SVD factor matrices. We demonstrate that our approach resolves the sampling difficulties encountered by standard Gibbs samplers for the common two-matrix factorization used in matrix completion.  More importantly, the geodesic Hamiltonian sampler allows for sampling in cases with more general likelihoods than the typical Gaussian likelihood and Gaussian prior assumptions adopted in most of the existing Bayesian matrix completion literature. We demonstrate an applications of our approach to fit the categorical data of a mice protein dataset and the MovieLens recommendation problem. Numerical examples demonstrate superior sampling performance, including better mixing and faster convergence to a stationary distribution. Moreover, they demonstrate improved accuracy on the two real-world benchmark problems we considered.
  
\smallskip
\noindent \textbf{Keywords.} Matrix Completion, Markov Chain Monte Carlo, Low-rank matrices  
\end{abstract}

%% file: intro.tex
\section{Introduction}\label{sec:intro}

We consider the problem of Bayesian low rank matrix completion. The matrix completion problem considers reconstructing a matrix $\mat{X} \in \reals^{m \times n}$ through indirect and noisy evaluations of a subset of its elements. A {\it low rank} matrix completion seeks to reduce the ill-posed nature of this problem by assuming that the underlying matrix has an approximately low rank factorization. A low rank factorization of a matrix  $\mat{X} \in \reals^{m \times n}$ matrix is defined by the couple $(\mat{A},\mat{B})$, where $\mat{A} \in \reals^{m \times r}$ and $\mat{B} \in \reals^{n \times r}$ with $r < \min(m, n)$ and
\begin{equation} \label{eq:low-rank}
  \mat{X} \approx \mat{A} \mat{B}^T.
\end{equation}
Recall that the best rank $r$ approximation $\mat{X}_r = \min_{\hat{\mat{X}}}\lVert \mat{X} - \hat{\mat{X}} \rVert_{F}^2$ is provided by the singular value decomposition
\begin{equation}\label{eq:svd}
  \mat{X}_{r} = \mat{U}_r \mat{\Lambda} \mat{V}_r^T,
\end{equation}
where $\mat{U}_r$ and $\mat{V}_r$ are the first $r$ columns of the left and right singular vectors of $\mat{X}$; $\mat{\Lambda}_r$ is an ordered set of the largest singular values; and where we can identify $\mat{A} = \mat{U}_r \sqrt{\mat{\Lambda}}$ and $\mat{B} = \mat{V}_r \sqrt{\mat{\Lambda}}.$

\paragraph{Bayesian low-rank matrix completion}
We consider the Bayesian setting aiming to identify distributions over the factors $\mat{A}$ and $\mat{B}$ to account for uncertainty stemming from an ability to uniquely identify the correct matrix. Data consists of pairs of element indices and values $\left\{\left(i,j\right), {y}_{ij} \right\}$ where $i \in [m]$ \rev{denotes the rows} and $j \in [n]$ \rev{denotes the columns}. We use the index set $\mathcal{I}$ to denote the set of indices corresponding to the observed data points and let $|\mathcal{I}|$ be its cardinality. The full data vector is denoted by $y$. We consider the following likelihood models: %
  a Gaussian likelihood 
\begin{align}\label{eq:gauss_like}
  y_{ij} &\sim \mathcal{N}(h(\mat{X}_{ij}), \sigma^2),
\end{align}
where $h$ is an observation function, e.g., identity or a soft-plus function to ensure positivity, and a Binomial likelihood
\begin{align}\label{eq:bino_like}
  y_{ij} &\sim \text{Bin}(k, h(\mat{X}_{ij})),  \quad
  h(\mat{X}_{ij}) =( 1 + \exp(-\mat{X}_{ij}) ) ^{-1}
\end{align}
where $k$ is the number of possible trials and $h(\mat{X}_{ij})$ denotes the success probability. \rev{Given the low rank factorization of Equation~\eqref{eq:low-rank}, here $\mat{X}_{ij}$ is obtained by multiplying the $i$th row of $\mat{A}$ with the $j$th row of $\mat{B}$.} The Binomial likelihood is applied in categorical settings where the categories might have a natural ordering. However, we emphasize that our proposed sampling methodology is more broadly applicable to other likelihood formulations. 

\begin{figure}
  \centering
  \includegraphics[width=0.7\textwidth]{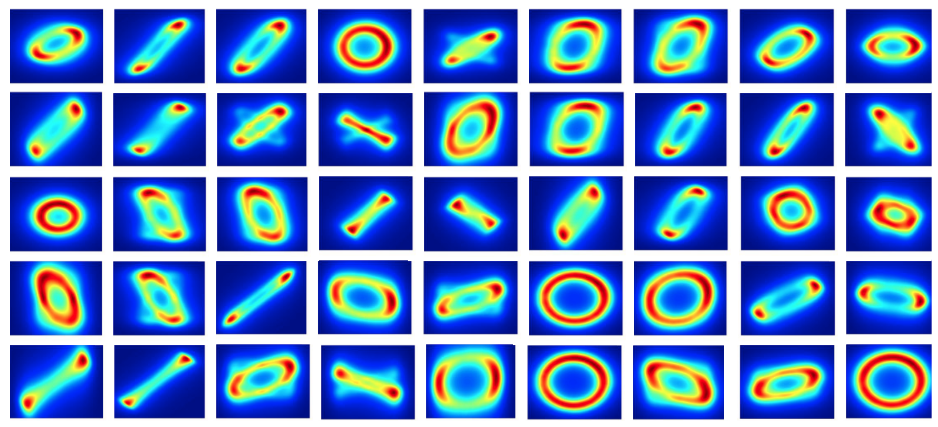}
  \caption{Examples of 2D marginals of the posteriors obtained using standard Gaussian priors for the inference of real-valued low-rank matrix factorizations~\cite{DeGorodetsky2020}.}
  \vspace{-15pt}
  \label{fig:hard_posteriors}
\end{figure}

\paragraph{Challenge}
One challenge to Bayesian matrix completion specifically, is the non-uniqueness of the low-rank representation. This challenge is pictorally represented in Figure~\ref{fig:hard_posteriors}, which demonstrates the geometry of posteriors that are prevalent in the problem. It is typically quite challenging to design a sampling scheme that is able to sample such posteriors, and even variational approaches are challenging to tune here. This challenge arises due to the following symmetry in the problem. A low-rank factorization can be obtained by an arbitrary linear transformation of the factors via
\begin{equation}
  \mat{X} = \mat{A}\mat{W} \mat{W}^{-1} \mat{B} = \hat{\mat{A}}\hat{\mat{B}}^T
\end{equation}
for any invertible $\mat{W} \in \reals^{r \times r}$. The precise ramifications of this symmetry in the Gaussian likelihood case was extensively documented in~\cite{DeGorodetsky2020}.

\paragraph{Contribution}
We develop an approach to address these issues in two ways: ({\romannumeral 1}) we alter the factorization format that constrains the factor matrices on Stiefel manifolds; and ({\romannumeral 2}) we develop a geodesic Markov Chain Monte Carlo approach that specifically target sampling distributions that lie on the manifold. Specifically, we propose to parameterize the low-rank matrix in the SVD format
\begin{equation}
  \mat{X} = \mat{U}\mat{S}\mat{V}^T, \quad \mat{U}^T\mat{U} = \mat{I}_r, \quad \mat{V}\mat{V}^T = \mat{I}_{r}
\end{equation}
where we seek to learn semi-orthogonal matrices $\mat{U}$ and $\mat{V}$ and a positive diagonal matrix $\mat{S}$. Within a Bayesian sampling context, we develop a geodesic Hamiltonian Monte Carlo (HMC)~\cite{byrne2013} within Gibbs algorithm to enable efficient sampling of the semi-orthogonal factor matrices.

This approach is closely related to the SVD models used in optimization-based methods~\cite{candes2010power} that solve a nuclear norm minimization problem
\begin{equation}
  \label{eq:matrix-recovery-nuclear}
  \Minimize_{\mat{X} \in \reals^{m \times n}} \quad \lVert \mat{X} \rVert _* \quad \text{subject to} \quad \lVert {y} - h(\Lambda(\mat{X})) \rVert \leq \delta,
\end{equation}
where $\Lambda$ is an operator that subselects the observed elements of $\mat{X}$ and flattens them into a vector of equal size to ${y}$. Moreover, an SVD parameterization is commonly used in this optimization procedure. From this perspective, the optimization problem can be seen as equivalently seeking a MAP estimate with a uniform prior over semi-orthogonal matrices. Prior work in Bayesian matrix completion has not used this equivalent prior to our knowledge. As a result, we provide the first Bayesian analogue of this extremely popular low-rank matrix characterization.

\paragraph{Related work} A large amount of work has addressed the low-rank Bayesian matrix completion problem in the variational setting. For example,
\cite{lim2007variational} and \cite{raiko2007principal} each apply variational Bayes approximations of this inference model to analyze the Netflix prize challenge \cite{bennett2007netflix} to great success.  Further, \cite{nakajima2011theoretical} and \cite{nakajima2013global} develop a theoretical framework to analyze the variational Bayes low-rank matrix factorization.
One of the issues with these existing approaches is the mean field-type approxiamtion that assumes independence of the factors. Sampling-based approaches target the full posterior and, while more expensive, have shown that this more complete characterization can outer-perform variational inference performance~\cite{salakhutdinov2008bayesian}. This approach was later extended to other settings \cite{chen2014stochastic, ahn2015large}, and have also been extended to the case of recovering low-rank tensor factorizations  \cite{rai2014scalable, zhao2015bayesian1, zhao2015bayesian2}.

The majority of these algorithms leverage the bilinear nature of the low-rank factorization to develop Gibbs samplers via conditional distributions for each factor. In this setting each factor is endowed with an independent Gaussian prior, and the variance of this prior is endowed with a hyperprior. This approach is mainly used to try to circumvent the sampling difficulties previously described, but comes at the cost of increased uncertainty. 
This approach uses a Gaussian log likelihood function
\begin{equation}
  \log P(y \mid  \mat{A}, \mat{B}) = -\frac{|\mathcal{I}|}{2} \log(2\pi) - |\mathcal{I}| \log(\sigma) - \frac{1}{2\sigma^2}\textstyle\sum_{(i,j)\in \mathcal{I}} \left(y_{ij} - \mat{A}_{i:} \mat{B}^T_{j:}\right)^2.
\end{equation}
The prior on the factors assigns i.i.d. zero mean Gaussians for entries of the matrices $\mat{A}$ and $\mat{B}$ and assumes the prior variance is unknown. This leads to a hierarchical prior, conditioned on the variance
\[
P(\mat{A}, \mat{B} \mid \delta) \propto ((m+n)r)^{-\frac12} \exp\left( -\frac{1}{2 \delta^2} \left( \textstyle\sum_{i=1,j=1}^{m,r} \mat{A}_{ij}^2 +  \textstyle\sum_{i=1,j=1}^{n,r}\mat{B}_{ij}^2 \right) \right),
\]
where $\delta$ is some unknown hyperparameter that controls the prior variance. We endow $\delta$ with a Gamma prior. Finally, Gibbs sampling is used to estimate both the factors $\mat{A}$ and $\mat{B}$, the prior variance $\delta$ and the likeihood standard deviation $\sigma$. 

\paragraph{Outline} This paper is structured as follows. Section 2 provides background on sampling on the Stiefel manifold and our proposed HMC within Gibbs algorithm. Section 3 provides numerical results that compare our proposed approach to Bayesian approach described above. The comparisons are made both on synthetic examples, to gain intuition, and on two real-world data sets that consider positivity constraints on the matrix elements and categorical valued matrices, respectively.

%% file: sampling.tex
\section{Sampling on low-rank manifolds}

In this section we first describe the geodesic Monte Carlo and then describe our Gibbs sampling procedure. The factor matrices we learn lie on the Stiefel manifold. Traditional Markov Chain Monte Carlo approaches like Hamiltonian Monte Carlo  or Riemannian-Manifold Hamiltonian Monte Carlo are not directly applicable to distributions for parameters that are on such manifolds embedded in the Euclidean space. Instead so-called geodesic Monte Carlo approaches propose modifications of HMC and related methods to directly apply to distributions in the embedded space~\cite{byrne2013}. 

\subsection{Operations on the Stiefel manifold}

We begin with description of operations on the Stiefel manifold. Let $\mathcal{M}$ denote an $m$ dimensional manifold that is embedded in a higher dimensional Euclidean space $\mathbb{R}^n.$ At every point $x \in \mathcal{X}$ there exists bijective mapping $q_x: \mathcal{M} \to \reals^{m}$ denoting a coordinate system on the manifold. The union of all possible coordinate systems across the manifold is called an atlas for a manifold. For our low-rank reconstruction purposes, we consider the Stiefel manifold of semi-orthogonal $n \times r$ matrices
$  \mathcal{M} = \{\mat{X} \in \reals^{n \times r}; \mat{X^TX} = \mat{I}_{r}\}.$

The main challenge is describing motion along the manifold since standard Euclidean vector-space addition does not apply. Instead, one must follow curves on the manifolds between points. A curve on the manifold over a time interval starting at some point $x \in \mathcal{M}$ is defined by $\gamma: [0,T] \to \mathcal{M}$, with $\gamma(0) = x$. At each point $x$, the tangent vector is an equivalence class of these curves\footnote{Two curves $\phi(t)$ and $\psi(t)$ are equivalent if $\phi(0) = \psi(0)$ and $\lim_{t\to0} \frac{\phi(t) - \psi(t)}{t} = 0$}, and the set of tangent vectors form a tangent vector space $TM_x \subset \mathbb{R}^n$~\cite{arnold2013}.

These tangent vectors are time derivatives of the curves $v = \dot{\gamma}(0) = \left. \frac{d}{dt} \gamma(t) \right|_{t=t0}.$ Every point on the Stiefel manifold statisfies $\mat{X}^T(t)\mat{X}(t) = \mat{I_{r}}$, so differentiation in time yields $\mat{\dot{X}}^T(t)\mat{X}(t) + \mat{X}^T(t)\mat{\dot{X}}(t) = 0.$ Therefore the tangent space is 
$  TM_{x} = \{\mat{Z} \in \reals^{n \times r}; \mat{Z}^T\mat{X} + \mat{X}^T\mat{Z} = \mat{0}\}.$

The geodesic Monte Carlo algorithm makes use of the space normal to the manifold. This normal space is the orthogonal complement to $TM_{x}$, and therefore a definition of an inner product is needed. \rev{For the Stiefel manifold this inner product is defined through the trace $\langle \mat{Y}, \mat{Z}\rangle_x = \mathrm{trace}(\mat{Y}^T\mat{Z}).$} One can then verify that for any $\mat{X} \in \mathcal{M}$ on the manifold, the orthogonal projection of $\mat{Z} \in \reals^{n \times r}$ onto the normal space is
\begin{equation}
\pi_{N_x}(\mat{Z})= \frac{1}{2}\mat{X}\left(\mat{X}^T\mat{Z} + \mat{Z}^T\mat{X}\right),
\end{equation}
and the orthogonal projection of $\mat{Z}$ onto the tangent space becomes  $(I - \pi_{N_x})(\mat{Z}) =  \mat{Z} - \pi_{N_x}(\mat{Z})$.
Finally we will require geodesics, the curve of shortest length, between two points on the manifold. Since the Steifel manifold is embedded in a Euclidean space, the geodesic can be defined by the fact that the acceleration vector at each point is normal to the manifold as long as the curve is traced with uniform speed. For Stiefel manifolds we are able to obtain the following closed form expression~\cite{edelman1998}
\begin{equation}   \label{eq:sol_on_geodesic}
  \left[\mat{X}(t), \mat{\dot{X}}(t)\right] = \left[\mat{X}(0), \mat{\dot{X}}(0)\right] \exp\left(
  \begin{bmatrix}
    \mat{A} & -\mat{S}(0) \\
    \mat{I} & \mat{A}
  \end{bmatrix}
  \right)
  \begin{bmatrix}
    \exp(-t\mat{A}) & 0 \\
    0 & \exp(-t\mat{A})
  \end{bmatrix}
\end{equation}
where $\mat{A} = \mat{X}^T(t)\mat{\dot{X}}(t)$ and $\mat{S}(t) = \mat{\dot{X}}^T(t)\mat{\dot{X}}(t).$
\subsection{Sampling on the Stiefel manifold}
While the Lebesgue measure is the reference measure for probability distributions in the Euclidean space, the Hausdorff measure can be used as the reference measure on manifolds~\cite{Diaconis2013}. It will be useful to move between an $m$-dimensional Hausdorff measure $\mathcal{H}^{m}$ and the $m$-dimensional Lebesgue measure $\lambda^{m}$. In the context of Riemannian manifolds, such as the Stiefel manifold, the relation between these measures is provided by the following formula~\cite{federer2014geometric}
\begin{equation}
  \mathcal{H}^{m}(dq) = \sqrt{|G(q)|} \lambda^{m}(dq),
  \label{eq:hausdorff}
\end{equation}
Where $G(q)$ is the Riemannian metric provided by the trace.
\rev{Using the metric $G(q)$ and the target density $\pi(q)$ with respect to the Lebesgue measure in the coordinate system provided by $q$, we can form the non-separable Hamiltonian function~\cite{MCMC:GiCal_2011}
\begin{equation}
  H(q, p) = - \log \pi(q) + \frac{1}{2}\log |G(q)| + \frac{1}{2}p^TG(q)^{-1}p,
  \label{eq:hamilton}
\end{equation}
where $\pi(q)$ is the target density with respect to the Lebesgue measure in the coordinate system provided by $q$, and $p$ is an auxiliary momentum variable following $\mathcal{N}(0, G(q))$. Note that $\exp(-H(q,p))$ defines a unnormalized joint probability density over the pair $(q,p)$, in which the marginal density over $q$ recovers the target density. As shown by the landmark paper of \cite{duane1987hybrid}, the Hamiltonian system 
\[
  \frac{dq_i}{dt} = \frac{\partial H}{\partial p_i}, \quad \frac{dp_i}{dt} = -\frac{\partial H}{\partial q_i},
\]
implicitly defines a time-reversible and one-to-one mapping $T$ from the state at time $t$, $(q(t), p(t))$, to the state at time $t + s$, $(q(t + s), p(t + s))$. This mapping $T$ leaves the target distribution $\pi(q)$ invariant. %
Thus, discretizing the Hamiltonian system using numerical integrators naturally leads MCMC proposal distributions that in general can reduce sample correlations in the resulting Markov chain. Moreover, using time-reversible and symplectic numerical integrators often yields simplified and more dimension-scalable MCMC proposals. See \cite{neal2011mcmc} and references therein. }

To define HMC on Riemannian manifold~\cite{MCMC:GiCal_2011}
, we can use the relation \eqref{eq:hausdorff} to write the Hamiltonian function~\eqref{eq:hamilton} with respect to the Hausdorff measure as
$ H(q,p) = - \log \pi_{\mathcal{H}}(q) + \frac{1}{2}p^TG(q)^{-1}p.$
The corresponding Hamiltonian system evolves the pair $(q,p)$ according to %
\begin{align}
  \frac{dq}{dt} = \frac{\partial H}{\partial p} = G(q)^{-1}p, \label{eq:pos_evolve}\qquad
  \frac{dp}{dt} = - \frac{\partial H}{\partial q} = \nabla_q \left[\log \pi_{\mathcal{H}}(q) - \frac{1}{2} p^T G(q)^{-1}p\right].
\end{align}
A key innovation of~\cite{byrne2013} is the introduction of a time-reversible and symplectic integrator that integrates this non-separable Hamiltonian system via splitting, $H(q,p) = H^{[1]}(q,p) + H^{[2]}(q,p)$ where $H^{[1]}(q,p) = -\log \pi_{\mathcal{H}}(q)$ and $H^{[2]}(q,p) = \frac{1}{2}p^TG(q)^{-1}p.$ The corresponding systems are
\begin{align}
  \frac{dq}{dt} & = 0, &
  \frac{dp}{dt} & =  \nabla_q \log \pi_{\mathcal{H}}(q) \label{eq:split1} \\
  \frac{dq}{dt} & = 2G^{-1}(q)p, &
  \frac{dp}{dt} & =  -p^T\frac{\partial }{\partial q} G^{-1}(q) p, \label{eq:split2}
\end{align}
respectively. The integrator first solves \eqref{eq:split1} for a timestep of $\epsilon / 2$, then solves \eqref{eq:split2} for a timestep of $\epsilon$, and finally propagates \eqref{eq:split1} again for a time-step of $\epsilon/2$. The solution of the first system~\eqref{eq:split1} is given simply by $(q(t),p(t)) = (q_0, p + t \nabla_q \log \pi_{\mathcal{H}}(q_0))$, and the solution to the second system\eqref{eq:split2} is given by the geodesic flow to enforce it on the manifold. 
If we denote $(\mat{X},\mat{V})$ as the embedded parameter and momentum, rather than $(q,p)$, the final HMC algorithm, made specific to the case of the Stiefel manifold, is provided in Algorithm~\ref{alg:stiefel_sample}. \rev{Note that the Metropolis-Hastings rejection rule is applied at the end of the time integration in Algorithm~\ref{alg:stiefel_sample} to endure the HMC generates Markov chains that have the target distribution $\pi(q)$ as the correct invariant distribution.}

\begin{algorithm}
  \caption{$\mathtt{hmc-stiefel}$: HMC step on the Stiefel Manifold~\cite{byrne2013}} \label{alg:stiefel_sample}
  \begin{algorithmic}[1]
    \Require timestep $\epsilon,$ number of steps $T$, current sample on Stiefel Manifold $\mat{X}$, target density $\pi_{\mathcal{H}}$
    \State $v \sim \mathcal{N}(0,\mat{I}_{nr})$ and  $\mat{V} \gets \textrm{reshape}(v, (n,r))$
    \State $H_{\text{current}} = \log \pi_{\mathcal{H}}(\mat{X}) - \frac12 v^Tv$\vspace{2pt}
    \State $\mat{V} \gets \mat{V} - \frac12\mat{X}(\mat{X}^T\mat{V} + \mat{V}^T\mat{X})$ and $v \gets \textrm{reshape}(\mat{V}, (nr))$\vspace{2pt}
    \For {$\tau = 1,\ldots, T$}\vspace{2pt}
    \State $v \gets v + \frac\epsilon2 \nabla_{x} \log \pi_{\mathcal{H}}( \mat{X}) $ and $\mat{V} \gets \textrm{reshape}(v, (n,r))$ 
    \Comment{Propagate velocity}
    \State $\mat{V} \gets \mat{V} - \frac{1}{2}\mat{X}(\mat{X}^T\mat{V} + \mat{V}^T\mat{X})$ 
    \Comment{Project back onto manifold}\vspace{2pt}
    \State $(\mat{X},\mat{V}) \gets$ solve Eq.~\eqref{eq:sol_on_geodesic} with initial condition $(\mat{X},\mat{V})$ for time $\epsilon$
    \Comment{Evolve on the geodesic}\vspace{2pt}
    \State $v \gets \textrm{reshape}(\mat{V}, (nr))$,  $v \gets v + \frac{\epsilon}{2} \nabla_{x} \log \pi_{\mathcal{H}}( \mat{X})$ and $\mat{V} \gets \textrm{reshape}(v, (n,r))$\vspace{2pt}
    \State $\mat{V} \gets \mat{V} - \frac{1}{2}\mat{X}(\mat{X}^T\mat{V} + \mat{V}^T\mat{X})$  and   $v \gets \textrm{reshape}(\mat{V}, (nr))$\vspace{2pt}
    \EndFor
    \State $H_{\textrm{next}} = \log \pi_{\mathcal{H}}(\mat{X}) - \frac{1}{2} v^Tv$\vspace{2pt}
    \State \algorithmicif\ {$\mathcal{U}[0,1] < \exp(H_{\textrm{next}} - H_{\textrm{current}})$}
    \algorithmicthen\ {$X_{\textrm(next)} \gets \mat{X}$}
  \end{algorithmic}
\end{algorithm}

\subsection{Application to low-rank matrix completion}

We utilize geodesic HMC within Gibbs to solve the matrix completion problem. The pseudocode is provided in Algorithm~\ref{alg:gibbs}. Here we see that the HMC on the Stiefel manifold is used to sample the left and right singular vectors, while a standard HMC approach is used for the singular values. In all cases we use the No-U-Turn-Sampler variant of HMC~\cite{hoffman2014no} to adaptively tune the number of steps.

\begin{algorithm}
  \caption{Geodesic HMC within Gibbs for matrix completion} \label{alg:gibbs}
  \begin{algorithmic}[1]
    \Require $\epsilon,$ timestep; $T$, number of steps; $\mat{X}$, current sample on the Stiefel Manifold; $N$ number of samples, $(\mat{U}^{(0)}, \mat{S}^{(0)}, \mat{V}^{(0)})$ initial samples; Conditional distributions $\pi_{U}(\mat{U} \mid , \mat{S}, \mat{V})$, $\pi_{S}(\mat{S} \mid \mat{U}, \mat{V})$, $\pi_{V}(\mat{V} \mid \mat{S}, \mat{U})$.
    \For {$k = 1,\ldots,N$}
    \State $\mat{U}^{(k)} \leftarrow \mathtt{hmc-stiefel}(\pi_{U}(\cdot \mid \mat{S}^{(k-1}), \mat{V}^{(k-1)}))$
    \State $\mat{V}^{(k)} \leftarrow \mathtt{hmc-stiefel}(\pi_{V}(\cdot \mid \mat{S}^{(k-1)}, \mat{U}^{(k)}))$
    \State $\mat{S}^{(k)} \leftarrow \mathtt{hmc}(\pi_{S}(\cdot \mid \mat{U}^{(k)},\mat{V}^{(k)}))$
    \EndFor
  \end{algorithmic}
\end{algorithm}

\paragraph{Prior}
For all the SVD based matrix-completion problems in this work, we consider independent prior distributions for the matrices $\mat{U},\mat{S}$ and $\mat{V}$. The priors of the left and right factors are defined as uniform in the Hausdorff measure, whereas the prior of the singular values are defined as the independent exponential distributions in a Euclidean space. This way, we have the prior
\begin{equation}
  \pi(\mat{U},\mat{S}, \mat{V}) = \pi_{\mathcal{H}}(\mat{U})P(\mat{S}) \pi_{\mathcal{H}}(\mat{V}) \propto \exp\left( - \lambda \textstyle\sum_{\ell = 1}^{r} \mat{S}_{\ell\ell}\right),
  \label{eq:prior}
\end{equation}
for some positive rate parameter $\lambda$. \rev{The exponential prior can be interpreted as the exponential of the negative nuclear norm of the underlying matrix $\mat{X}$, i.e., $\lambda \|\mat{X}\|_\ast = \lambda \sum_{\ell = 1}^{r} \mat{S}_{\ell\ell}$, which is a convex envelope of $\text{rank}(\mat{X})$ used in the optimization literature  to promote low-rankness in matrix estimation problems \cite{candes2010power,mazumder2010spectral,hastie2015matrix}. Thus, the exponential prior can be viewed as a rank revealing prior. We provide further justifications and numerical illustrations in Section C of the supplementary material.}

In the rest of this section, we describe three likelihood models used for the numerical experiments: (1) real-valued data, (2) positive real-valued data for the mice protein expression example, and (3) positive integer-valued rating data for the movie recommendation.

\paragraph{Model SVD:} In this model, we treat the observed data as real-valued, and consider the Gaussian likelihood arising from the model
\begin{equation}\label{eq:gauss_data}
  y_{ij} = \mat{X}_{ij} + \xi =  \mat{U}_{i:}\mat{S}\mat{V}_{j:}^T + \xi, 
\end{equation}
where $\xi \sim \mathcal{N}(0, \sigma^2).$ The log likelihood becomes
\begin{equation*}
  \log P(y \mid  \mat{U}, \mat{S}, \mat{V}) = -\frac{|\mathcal{I}|}{2} \log(2\pi) - |\mathcal{I}|\log(\sigma) - \frac{1}{2\sigma^2}\textstyle\sum_{(i,j)\in\mathcal{I}} \left (y_{ij} - \mat{U}_{i:} \mat{S} \mat{V}^T_{j:}\right)^2.
\end{equation*}
\paragraph{Model S-SVD:} In this model, we treat the observed data as positive real-valued, and consider the Gaussian likelihood arising from the model
\begin{equation}\label{eq:soft_data}
  y_{{\bf i}} = h(\mat{X}_{ij}) + \xi =  h(\mat{U}_{i:}\mat{S}\mat{V}_{j:}^T) + \xi, \quad h(z) = \log(1 + \exp(z)),
\end{equation}
where $\xi \sim \mathcal{N}(0, \sigma^2).$ The log likelihood becomes
\begin{equation*}
  \log P(y \mid  \mat{U}, \mat{S}, \mat{V}) = -\frac{|\mathcal{I}|}{2} \log(2\pi) - |\mathcal{I}|\log(\sigma) - \frac{1}{2\sigma^2}\textstyle\sum_{(i,j)\in\mathcal{I}} \left (y_{ij} - h(\mat{U}_{i:} \mat{S} \mat{V}^T_{j:})\right)^2.
\end{equation*}
\rev{For Model SVD and Model S-SVD, the maximum {\it a-posteriori} (MAP) of the posterior becomes equivalent to a penalized version of the nuclear norm minimization problem~\eqref{eq:matrix-recovery-nuclear}. This equivalence is obtained by taking the log of the prior~\eqref{eq:prior} and adding either of the log likelihoods described above.}

\paragraph{Model B-SVD:} In this model, we treat the observed data as positive integer-valued, and consider the Binomial likelihood, 
\begin{equation}\label{eq:bino_data}
  y_{ij} \sim \text{Bin}(k,h(\mat{X}_{ij})),  \quad h(\mat{X}_{ij}) = ( 1 + \exp(-\mat{X}_{ij}) ) ^{-1} = ( 1 + \exp(-\mat{U}_{i:} \mat{S} \mat{V}^T_{j:}) ) ^{-1}.
\end{equation}
This yields the log likelihood
\begin{equation*}
 \log P(y \mid  \mat{U}, \mat{S}, \mat{V}) = \textstyle\sum_{(i,j)\in\mathcal{I}} \left(\log {k \choose y_{ij}} + y_{ij} \log h(\mat{U}_{i:} \mat{S} \mat{V}^T_{j:}) + \left(k - y_{ij}\right) \log \left(1 - h(\mat{U}_{i:} \mat{S} \mat{V}^T_{j:})\right)\right).
 \end{equation*}

%% file: examples.tex
\section{Numerical examples}~\label{sec:examples}

We now demonstrate the efficacy of the proposed SVD-based models and the geodesic HMC sampling methods on three examples, including partially observed synthetic matrices, the mice protein expression data set \cite{higuera2015self}, and the MovieLens data set \cite{harper2015movielens}. 

\paragraph{Problem setups}
Example 1 includes the following synthetic test cases that aim to benchmark the proposed methods.
\begin{itemize}[align=left]
\item[Case \#1:] We construct a true matrix \( \mat{X}_\text{true} = \mat{A} \mat{B}^T \), where entries of $\mat{A}, \mat{B}$ are drawn from i.i.d. standard Gaussian, and then partially observed data $y_{ij} = \mat{X}_{ij}$ are given by randomly selecting a subset of matrix indices. 

\item[Case \#2:] We construct a true positive matrix \( \mat{X}_\text{true} = \mat{A} \mat{B}^T \), where entries of $\mat{A}$ are drawn from i.i.d. $\text{Exp}(\cdot, 1)$ and entries of $\mat{B}$ are drawn from i.i.d. $\text{Uniform}(\cdot, [0,1])$. Then partially observed data $y_{ij} = \mat{X}_{ij}$ are given by randomly selecting a subset of matrix indices.

\item[Case \#3:] We construct a true matrix \( \mat{X}_\text{true} = \mat{A} \mat{B}^T \), where entries of $\mat{A}, \mat{B}$ are drawn from i.i.d. standard Gaussian. Then partially observed data are drawn from the Binomial model, i.e., $y_{ij} \sim \text{Bin}(k,( 1 + \exp(-\mat{X}_{ij}) ) ^{-1})$, where the indices are randomly chosen. 
\end{itemize}
For each of the cases, we use $100\times 60$ matrices with rank $10$. We conduct two sets of experiments that respectively select $10\%$ and $40\%$ of matrix entries as training data. The cross validation data sets are chosen as $40\%$ of matrix entries that do not overlap with the training data sets. 

Example 2 is the mice protein expression data set, which consists of $77$ protein expressions, measured in terms of nuclear fractions, from $1080$ mice specimens. The $1080 \times 77$ matrix contains positive real-valued entries. This data set is openly available at the UCI Machine Learning Repository \cite{Dua:2019}. We use rank $20$ in the training. Here we also conduct two sets of experiments that respectively select $10\%$ and $40\%$ of matrix entries at random as training data. The cross validation data sets are chosen as $40\%$ of matrix entries at random that do not overlap with the training data sets.

Example 3 uses the \texttt{ml-latest-small} of the MovieLens data set, which contains $100836$ non-negative ratings across $9742$ movies rated by 610 users. This gives a $610 \times 9724$ matrix with $100836$ partially observed rating data\footnote{The data set is openly available at \url{https://files.grouplens.org/datasets/movielens/}}. For the rating data, each of the observed data entries $y_{ij}$ is a categorical variable, which takes value from a finite sequence in the range $[0,5]$ with increment $0.5$, i.e., $y_{ij} \in \{0, 0.5, 1, \ldots, 4, 4.5, 5\}$. To apply the Binomial likelihood, we scale the rating data by a factor of two, so they become integer-valued in the ranging $[0,10]$ with increment $1$.  We use rank $20$ in the training. In this example, we randomly select $80\%$ of the data in the data set as the training data, and use the rest of $20\%$ as cross validation data. 

Here Case \#1 of the synthetic example aims to benchmark the expression power and the sampling efficiency of the SVD model in comparison with the $\mat{A}\mat{B}^T$ model. For this reason, the ground truth is generated from the prior distribution of the $\mat{A}\mat{B}^T$ model. The synthetic cases 2 and 3 mimic the behavior of the mice protein expression data set \cite{higuera2015self} and the MovieLens data set \cite{harper2015movielens}, respectively.

\begin{table}[h]
\caption{Summary of sampling performance and prediction accuracy. The model $\mat{A}\mat{B}^T$ is sampled by the Gibbs sampling. The SVD-based models are sampled by geodesic HMC. For table entries shown with $\nicefrac{}{}$, the left and the right represent results for the $10\%$ and $40\%$ data sampling rates, respectively.}
\label{tab:overview}
\footnotesize
\begin{center}
\renewcommand{\arraystretch}{1.3}
\hspace{-5em}
\begin{tabular}{l|c|c|c|c|c}
\toprule
& \multicolumn{2}{c|}{MCMC mixing} & \multicolumn{3}{c}{$[1\%, 50\%, 99\%]$ quantiles of MADs}\\
\cline{2-6}
& Gibbs & HMC &$\mat{A}\mat{B}^T$& SVD & $\ast$-SVD \\
\hline
 Case \#1  & {\ding{55}} & {\ding{51}} & $\nicefrac{[0.04, 2.3, 9.8]}{[0.04, 2.3, 10.0]}$ & $\nicefrac{[0.04, 2.3, 9.8]}{[0.04, 2.3, 10.0]}$ & - \\
\cline{2-6}
 Case \#2  & {\ding{55}} & {\ding{51}} & $\nicefrac{[0.20, 6.3, 24]}{[0.67, 6.1, 13]}$ & $\nicefrac{[1.9, 6.2, 13]}{[0.52, 6.1, 13]}$ & $\nicefrac{[1.9, 6.2, 13]}{0.58, 6.1, 13]}$\\
\cline{2-6}
 Case \#3  & {\ding{55}} & {\ding{51}} & $\nicefrac{[0.15, 6.3, 21.8]}{[0.69, 6.1, 13.4]}$ & $\nicefrac{[0.15, 5.5, 10.5]}{[0.24, 5.4, 10.4]}$ & $\nicefrac{[0, 5, 10]}{[0, 5, 11]}$ \\
\hline
 Mice  & {\ding{55}} & {\ding{51}} & $\nicefrac{[0, 0.33, 2.6]}{[0, 0.22, 2.4]}$ &  $\nicefrac{[0, 0.33, 2.6]}{[0.02, 0.22, 2.4]}$ & $\nicefrac{[0, 0.36, 2.1]}{[0.02, 0.42, 1.8]}$ \\
\hline
MovieLens & {\ding{55}} & {\ding{51}} & \scriptsize $[0.02, 1.2, 8.0]$ & \scriptsize $[0.03, 1.6, 5.0]$ & \scriptsize $[0, 1, 5]$  \\
\bottomrule
\end{tabular}
\end{center}
\end{table}

A summary of the sampling performance and the prediction accuracy of various methods is provided in Table \ref{tab:overview}.  The table specifies if the sampling method converges to the posterior distribution within 10000 iterations including 2000 iterations of burnin, and also reports the estimated median absolute deviations (MAD) on the cross validation data sets and the standard deviations of the MADs. \rev{Since both the HMC and the Gibbs sampling algorithms discussed here are asymptotically converging methods when they are simulated for an infinite number of iterations \cite{liu2001monte,robert1999monte}, here we only determine the converging behavior with finite iterations using sample traces. For Markov chains with reasonable mixing, we then compute the autocorrelation time to benchmark the rate of convergence \cite{liu2001monte}.}
More details about the sampling performance and the prediction accuracy of various models applied to the abovementioned examples are provided in the rest of this section.

\begin{figure}[h]
  \centering
  \includegraphics[height=0.2\textwidth, trim = {6em, 0, 6em, 0},  clip]{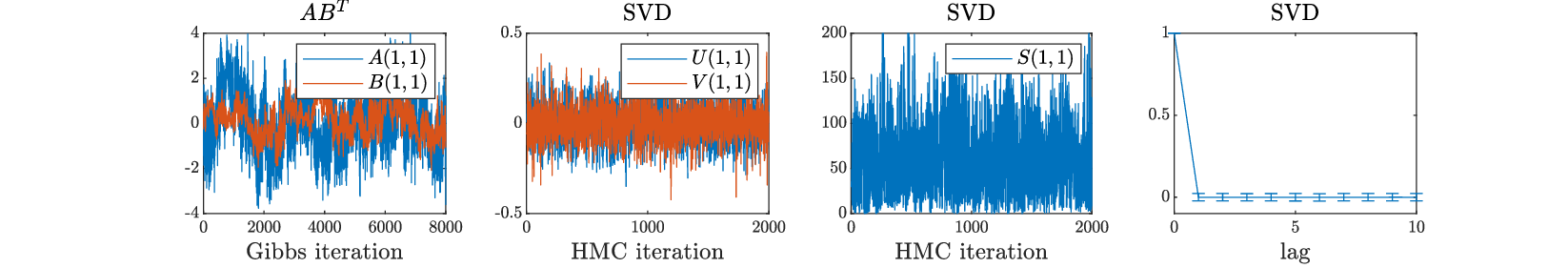}
   \caption{One random experiment of Case \#1 of the synthetic example, with $10\%$ data sampling rate. From left to right: MCMC traces produced by $\mat{A}\mat{B}^T$ and SVD, and the average autocorrelation times $\pm$ standard deviations produced by HMC. Error bars are obtained using 30 random experiments.}\label{fig:trace_case1}
\end{figure}

\paragraph{Sampling performance} Overall, the Gibbs sampling on the $\mat{A}\mat{B}^T$ model demonstrates a poor mixing behavior in our examples. The Markov chains often do not converge to the stationary distribution after many iterations. In comparison, all Markov chains generated by the HMC on the SVD-based model demonstrate a superior mixing property. For brevity, we only report the mixing of Markov chains in Case \#1 of the synthetic example here, in which the ground truth is generated by the $\mat{A}\mat{B}^T$ model. As shown in Figure \ref{fig:trace_case1}, after $8 \times 10^3$ sweeps of Gibbs updates on the parameters $(\mat{A}, \mat{B}, \delta, \sigma)$, the traces of the Markov chain cannot reach the stationary distribution. In comparison, the traces of the geodesic HMC can reach the stationary distribution very quickly. To further demonstrate the sampling efficiency of HMC, we reported the estimated average autocorrelation time over all parameters in Figure \ref{fig:trace_case1} (right), which shows that the autocorrelation of the Markov chains generated by HMC decreases to zero after one iteration. %
We observe similar mixing behavior in all other test cases in the synthetic example and also in examples 2 and 3. See the appendices for the mixing diagnostics of the rest of examples.

We observe that each iteration of the geodesic HMC requires more computational effort compared to one sweep of the Gibbs. For Case \#1 of the synthetic example, the CPU time per iteration of the geodesic HMC is about $5$ times of that of the Gibbs. For positive matrices (Case \#2 and the mice protein data), the CPU time per iteration of the geodesic HMC is about $15-30$ times of that of the Gibbs. For the rating data (Case \#3 and the MovieLens), the CPU time per iteration of the geodesic HMC is about $2-8$ times of that of the Gibbs. However, the rather high computational cost of the geodesic HMC clearly yields a superior sampling performance.

\begin{figure}
  \centering
  \includegraphics[height=0.2\textwidth, trim = {6em, 0, 6em, 0},  clip]{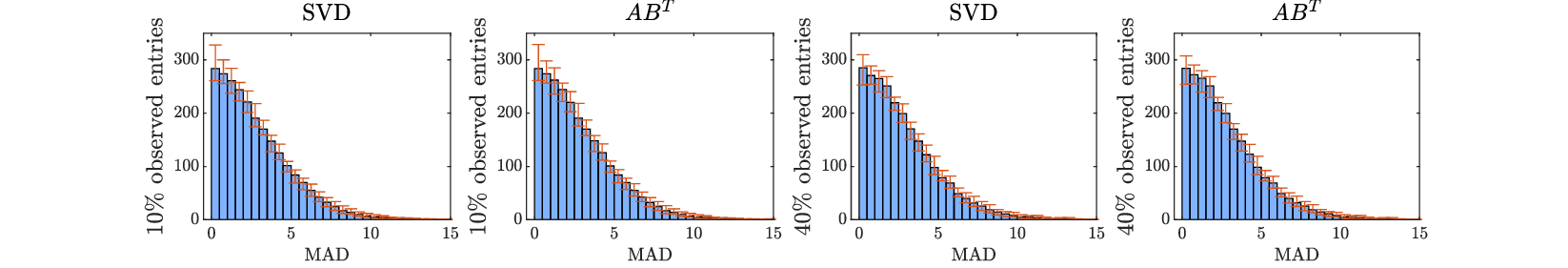}
  \caption{Case \#1 of the synthetic example. Histograms of prediction MADs estimated using different approaches. The reported error bars are obtained using 30 random experiments. The SVD model is sampled by HMC, while the $\mat{AB}^T$ model is sampled by Gibbs.\label{fig:mad_case1}}
\end{figure}

\paragraph{Prediction accuracy} For all the examples, we report the histograms of the posterior MADs (compared to the cross validation data sets) generated by various models. For all the synthetic test cases, the uncertainty intervals on the histogram are computed using 30 random experiments. For examples 2 and 3, the uncertainty intervals are computed using 10 random experiments.

Figure \ref{fig:mad_case1} shows the MADs for Case \#1 of the synthetic example. For both data sampling rates, the SVD model and the $\mat{A}\mat{B}^T$ model produce comparable MADs in this case.

\begin{figure}

  \begin{subfigure}[b]{0.5\linewidth}
  \includegraphics[width=\textwidth, trim = {0 1.2em 0em 0}, clip]{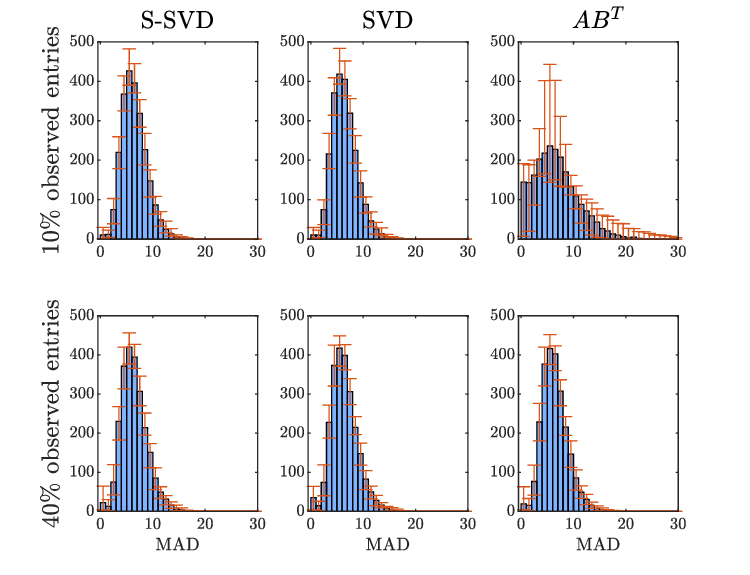}
    \caption{}
  \end{subfigure}\hspace{-6pt}
  \begin{subfigure}[b]{0.5\linewidth}
  \includegraphics[width=\textwidth, trim = {0 1.2em 0em 0}, clip]{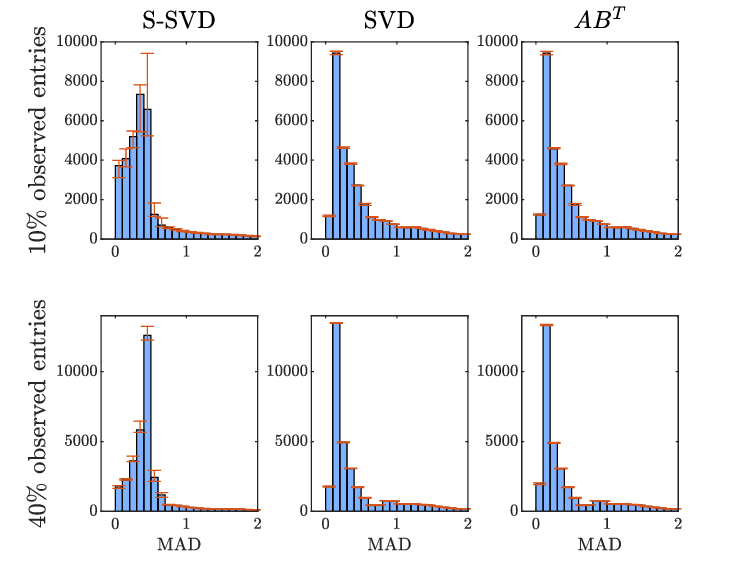}
    \caption{}
  \end{subfigure}%

  \caption{Histograms of prediction MADs for positive matrices. (a): Case \#2. (b): mice data. For both (a) and (b), the top row and bottom row correspond to $10\%$ and $40\%$ data sampling rates, respectively; from the left column to the right column we have the S-SVD, SVD, and $\mat{A}\mat{B}^T$ models. The S-SVD model generally achieves better performance with smaller tails than the $\mat{AB}^T$ model. \label{fig:mad_case2}}
\end{figure}

Figure \ref{fig:mad_case2} shows the MADs for Case \#2 of the synthetic example and the mice protein data set. Both examples here involve matrices with positive entries. We apply the S-SVD model (which preserves the positivity), the SVD model and the $\mat{A}\mat{B}^T$ model. For Case \#2, Figure \ref{fig:mad_case2} (a) shows that with a high data sampling rate ($40\%$), all three models produce comparable MADs. However, with a low data sampling rate ($10\%$), the S-SVD model and the SVD model can still generate similar results compared to the $40\%$ data sampling rate case, whereas the MADs of the $\mat{A}\mat{B}^T$ model exhibits some heavy-tail behavior. For the mice protein data set, Figure \ref{fig:mad_case2} (b) shows that the S-SVD model outperforms the SVD model and the $\mat{A}\mat{B}^T$ model in the low data sampling rate case, while all three models have similar performance in the high data sampling rate case (note that the S-SVD model has the smallest uncertainty interval). Overall, the S-SVD model demonstrates a clear advantage. 

\begin{figure}
  \begin{subfigure}[b]{0.5\linewidth}
  \includegraphics[width=\textwidth, trim = {0 1.2em 0em 0}, clip]{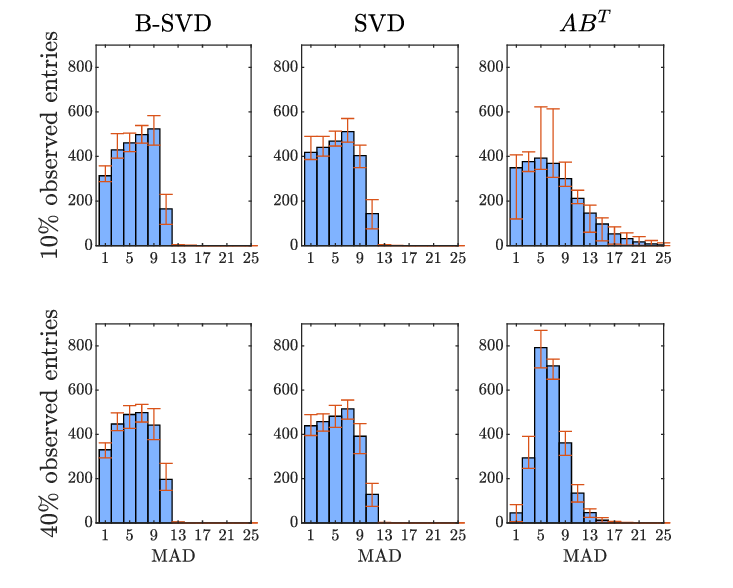}
    \caption{}
  \end{subfigure}%
  \begin{subfigure}[b]{0.5\linewidth}
  \includegraphics[width=\textwidth, trim = {0 .3em 0em 0}, clip]{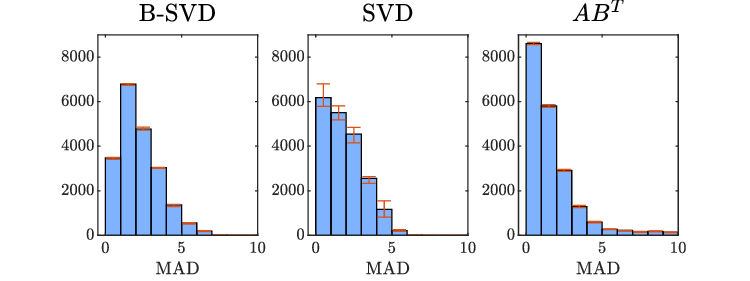}
    \caption{}
  \end{subfigure}%
  \caption{Histograms of prediction MADs for rating data. (a): Case \#3, where the top row and bottom row correspond to $10\%$ and $40\%$ data sampling rates, respectively. (b): MovieLens. For both (a) and (b), from the left column to the right column we have the B-SVD, SVD, and $\mat{A}\mat{B}^T$ models.}  \label{fig:mad_case3}
\end{figure}

Figure \ref{fig:mad_case3} shows the MADs for Case \#3 of the synthetic example and the MovieLens data set. Both examples here involve rating data. We apply the B-SVD model (which uses the Binomial likelihood), the SVD model and the $\mat{A}\mat{B}^T$ model. For Case \#3, Figure \ref{fig:mad_case3} (a) shows that for both data sampling rates, the two SVD-based models generate comparable results and are both more accurate compared to that of the $\mat{A}\mat{B}^T$ model.

For the MovieLens case, Figure \ref{fig:mad_case3} (b) shows that the S-SVD model outperforms the SVD model and the $\mat{A}\mat{B}^T$ model in the low data sampling rate case. All three models have similar performance in the high data sampling rate case. Furthermore, the S-SVD model has the smallest uncertainty interval. Overall, the S-SVD model demonstrates a clear advantage in the prediction accuracy.

%% file: appendix.tex
\section{Vectorization of conditional posteriors of SVD-based models}\label{sec:svd_conditionals}

In this section we derive the conditional posteriors of $\mat{U}$, $\mat{S}$, and $\mat{V}$ in the singular value decomposition model with a Gaussian likelihood. The same procedure can be undertaken for the posteriors associated with all the proposed likelihood models.

As a starting point, we let $\gamma = 1/\sigma^2$ to denote the precision parameter of the Gaussian observation noise and assign it a conjugate Gamma prior $\Gamma( \alpha_{\sigma}, \beta_{\sigma}) $ with $\alpha_{\sigma} = \beta_{\sigma} = 10^{-4}$. We have the full posterior distribution
\begin{equation}\label{eq:full_post}
  P(\mat{U}, \mat{S}, \mat{V}, \gamma | y) \propto \gamma^{\frac{|\mathcal{I}|}2} \exp\left( - \frac{\gamma}{2}\sum_{(i,j)\in\mathcal{I}} \left (y_{ij} - \mat{U}_{i:} \mat{S} \mat{V}^T_{j:}\right)^2 \right) \pi_{\mathcal{H}}(\mat{U})P(\mat{S}) \gamma ^{\alpha_\sigma - 1} \exp( - \beta_\sigma \gamma ),
\end{equation}
where $\pi_{\mathcal{H}}(\mat{U})$ and $\pi_{\mathcal{H}}(\mat{V})$ are uniform prior densities in the Hausdorff measure and $P(\mat{S})$ is some prior density for the singular values. 
Given $(\mat{S},\mat{V},\gamma)$, the conditional posterior distribution for $\mat{U}$ is
    \begin{align}
      \pi_{U}\left(\mat{U} \mid y, \mat{S}, \mat{V}, \gamma\right) &\propto \exp\left[ - \frac{\gamma}{2}\sum_{(i,j)\in \mathcal{I}} \left (y_{ij} - \mat{U}_{i:} \mat{S} \mat{V}^T_{j:}\right)^2\right] 
      = \exp\left[-  \frac{\gamma}{2} \sum_{(i,j)\in \mathcal{I}} \left (y_{ij} - \mat{U}_{i:} \mat{B}_j\right)^2 \right]
      \label{eq:post}
    \end{align}
    where $\mat{B}_j = \mat{S}\mat{V}^T_{j:}$. In the implementation it is easier to work directly with $\mat{U}$ rather than repeated indexing into its rows. To this end, it is convenient to reorder $y$ according to the row indices so that we have
    \begin{align}\label{eq:vec_y}
      y  = \begin{bmatrix}
        \mat{y}_1 \\
        \vdots \\
        \mat{y}_m
      \end{bmatrix}
      &=
      \begin{bmatrix}
        \tilde{\mat{B}}_1 &  &  \\
         & \ddots &  \\
         &  & \tilde{\mat{B}}_{m}
      \end{bmatrix}
      \begin{bmatrix}
        \mat{U}^T_{1:} \\
        \vdots \\
        \mat{U}^T_{m:}
      \end{bmatrix}
      + \mat{\xi} =
     \mat{B} \ \mathtt{reshape}(\mat{U}, (mr)) + \mat{\xi},
    \end{align}
where $m_k$ is the number of observed entries in the $k$-th row of the data matrix so that $\mat{y}_k \in \reals^{m_{k}}$; $\tilde{\mat{B}}_k \in \reals^{m_{k} \times r}$ is a matrix consisting of vertically concatenated blocks of $\mat{B}_j$ corresponding to the columns $j$ of the observed entries in row $k$; $\mat{U}_{k:} \in \reals^{r}$ is each row of $\mat{U}$; and $\mat{B}$ is a block-diagonal matrix. Together with the uniform prior in the Hausdorff measure, this results in the following conditional posterior
    \begin{equation} \label{eq:cond_post_U}
      \pi_{U}(\mat{U} \mid y, \mat{S}, \mat{V}, \gamma) \propto \exp\left[ - \frac{\gamma}{2} \lVert y - \mat{B}\  \texttt{reshape}(\mat{U}, (mr)) \rVert^2 \right].
    \end{equation}
   This density does not correspond to that of a  Gaussian distribution over Euclidean space because the matrix must satisfy orthogonality properties. Thus there is no closed-form approach for sampling from this posterior, and some MCMC method is needed. The conditional posterior for $\mat{V}$ can be obtained in a similar way.

    Next we consider the conditional posterior of the singular values lying along the diagonal of the matrix $\mat{S}$. We can assign any prior with positive support for the diagonal elements of this matrix. In this paper we choose an exponential prior $P(\mat{S}) \propto \exp( - \lambda \sum_{\ell=1}^{r} \mat{S}_{\ell \ell}) $, which draws an analogy with the nuclear norm regularization. Then, Equation~\eqref{eq:full_post} leads to the conditional posterior
    \begin{align}
      \pi_{S}\left(\mat{S} \mid y, \mat{U}, \mat{V}, \gamma \right) \propto  \exp\left[ - \frac{\gamma}{2}\sum_{(i,j)\in \mathcal{I}} \left (y_{ij} - \mat{U}_{i:} \mat{S} \mat{V}^T_{j:}\right)^2 - \lambda \sum_{\ell=1}^{r} \mat{S}_{\ell \ell}\right]
    \end{align}
    where $\lambda$ is the hyperparameters of the exponential distribution. To further simplify the computation, we can also express the above conditional posterior in the form of    
    \begin{align}\label{eq:cond_post_S}
      \pi_{S}\left(\mat{S} \mid y, \mat{U}, \mat{V}, \gamma \right) \propto \exp\left[ - \frac{\gamma}{2} \lVert y - \mat{M}\  \texttt{diag}(\mat{S}) \rVert^2  - \lambda \sum_{\ell=1}^{r} \mat{S}_{\ell \ell} \right],
    \end{align}
    where the vector $y \in \reals^{|\mathcal{I}|}$ is the same as the one defined in \eqref{eq:vec_y}; $\texttt{diag}(\mat{S})$ maps the diagonal entries of the matrix $\mat{S} \in \reals^{r \times r}$ into a vector in $\reals^r$ consisting of all the singular values; and the matrix $\mat{M} \in \reals^{|\mathcal{I}| \times r}$ is given as
    \[
		\mat{M} = \begin{bmatrix}
        \mat{U}_{i_1:} \circ \mat{V}_{j_1:} \\
        \vdots \\
        \mat{U}_{i_k:} \circ \mat{V}_{j_k:} \\
        \vdots
      \end{bmatrix} ,
    \]
    where $\circ$ denotes the Hadamard product. 

    The computational complexity of evaluating the conditional densities \eqref{eq:cond_post_U} and \eqref{eq:cond_post_S} and their gradient are governed by the cost of matrix-vector-products with $\mat{B}$ and $\mat{M}$, which are both $O(|\mathcal{I}| r)$.

    Finally, the conditional posterior of the precision parameter $\gamma$ takes the form
\begin{equation}\label{eq:full_post_2}
  P(\gamma | y, \mat{U}, \mat{S}, \mat{V}) \propto \gamma^{\frac{|\mathcal{I}|}2 + \alpha_\sigma - 1} \exp\left[ - \left( \frac{1}{2}\sum_{(i,j)\in\mathcal{I}} \left (y_{ij} - \mat{U}_{i:} \mat{S} \mat{V}^T_{j:}\right)^2 + \beta_\sigma\right) \gamma \right],
\end{equation}
which is another Gamma distribution that can be sampled directly.

\section{Gibbs $\mat{A}\mat{B}^T$ model}

In this section we provide the pseudocode of the Gibbs sampler used for sampling the $\mat{AB}^T$ model. Recall that the unknown parameters are the low-rank factors $\mat{A}, \mat{B}$, the measurement noise variance  $\sigma^2$, and the prior variance of the factors $\delta^2$. Similar to the SVD model, we use the precision parameters $\gamma = 1/\sigma^2$ and $\tau = 1/\delta^2$ and assign conjugate Gamma distributions $\Gamma(\alpha_\sigma, \beta_\sigma)$ and $\Gamma(\alpha_\delta, \beta_\delta)$, respectively. Here all the rate and shape parameters of the Gamma prior is set to $10^{-4}$. Together these facts lead to the following full posterior
\begin{align}
  P(\mat{A}, \mat{B}, \gamma, \tau \mid y) \propto &  \gamma^{\frac{|\mathcal{I}|}2} \exp\left[ - \frac{\gamma}{2}\sum_{(i,j)\in \mathcal{I}} \left (y_{ij} - \mat{A}_{i:} \mat{B}^T_{j:}\right)^2 \right] \gamma ^{\alpha_\delta - 1} \exp( - \beta_\delta \gamma )  \times \nonumber \\
  & \tau^{\frac{(m+n)r}2}  \exp\left[ - \frac{\tau}{2} \left( \sum_{i=1,j=1}^{m,r} \mat{A}_{ij}^2 +  \sum_{i=1,j=1}^{n,r}\mat{B}_{ij}^2 \right) \right]  \tau ^{\alpha_\delta - 1} \exp( - \beta_\delta \tau )
\end{align}

The conditional distributions for $\mat{A}$ and $\mat{B}$ are both Gaussian distributions and so samples can be drawn using Gibbs sampling. However, directly sampling those Gaussian distributions needs to factorize rather large covariance matrices, e.g., a matrix with dimension $mr \times mr$ for $\mat{A}$. This can have superlinear computational complexity, for instance, cubic complexity if directly applying the Cholesky factorization. 
One way to reduce the computational cost is to apply Gibbs sampling to each of $\mat{A}$ and $\mat{B}$ row-by-row. Denoting $\mat{A}_{i:}$ the $i$-th row of $\mat{A}$ and $\mat{A}_{-i:}$ the matrix containing remaining rows of $\mat{A}$, the conditional posterior of $\mat{A}_{i:}$ given $(\mat{A}_{-i:}, \mat{B}, \gamma, \tau)$ is
\[
  P(\mat{A}_i |y , \mat{A}_{-i:} \mat{B}, \gamma, \tau) \propto 
  \left\{ \begin{array}{ll} 
  \exp\left[ - \frac{\gamma}{2}\sum_{j \in \mathcal{J}_i} \left (y_{ij} - \mat{A}_{i:} \mat{B}^T_{j:}\right)^2 - \frac{\tau}{2} \left( \sum_{j=1}^r \mat{A}_{ij}^2 \right) \right] & i \in \mathcal{I}\\  
  \exp\left[ - \frac{\tau}{2} \left( \sum_{j=1}^{r} \mat{A}_{ij}^2 \right) \right] & i \notin \mathcal{I}
  \end{array} \right. , 
\]
where $\mathcal{J}_i$ is the index set containing all the column indices such that $(i,j) \in \mathcal{I}$ for a given row index $i$. Note the second conditional refers to the situation where row $i$ of the matrix is not observed at all, in which case sampling proceeds from the prior. The conditional posterior of $\mat{B}_{i:}$ can be derived in a similar way. Each of the conditional Gaussian distributions has dimension $r$, and thus can be computationally much cheaper to simulate. 
The pseudocode is provided in Algorithm~\ref{alg:gibbs_ab}. 

\begin{algorithm}
  \caption{Gibbs sampler for the $\mat{AB}^T$ model} \label{alg:gibbs_ab}
  \begin{algorithmic}[1]
    \Require $N$ number of samples, $(\mat{A}^{(0)}, \mat{B}^{(0)}, \gamma^{(0)}, \tau^{(0)})$ initial samples; 
    \For {$k = 1,\ldots,N$}
    \State $\gamma^{(k)} \leftarrow \mathtt{sample\_gamma}(P(\gamma | y, \mat{A}^{(k-1)}, \mat{B}^{(k-1)}, \tau^{(k-1)} )) $
    \State $\tau^{(k)} \leftarrow \mathtt{sample\_gamma}(P(\tau | y, \mat{A}^{(k-1)}, \mat{B}^{(k-1)}, \gamma^{(k)} ))  $
    \State $\mat{A}^{(k)} \leftarrow \mat{A}^{(k-1)}$
    \For {$i = 1,\ldots,m$}
		\State $\mat{A}^{(k)} \leftarrow \mathtt{sample\_gaussian}(P(\mat{A}_{i:} | y, \mat{A}^{(k)}_{-i:}, \mat{B}^{(k-1)}, \gamma^{(k)}, \tau^{(k)} ) )$
    \EndFor
    \State $\mat{B}^{(k)} \leftarrow \mat{B}^{(k-1)}$
    \For {$j = 1,\ldots,n$}
		\State $\mat{B}_{j:}^{(k)} \leftarrow \mathtt{sample\_gaussian}(P(\mat{B}_{j:} | y, \mat{B}^{(k)}_{-j:}, \mat{A}^{(k)}, \gamma^{(k)}, \tau^{(k)} ) )$
    \EndFor
    \EndFor
  \end{algorithmic}
\end{algorithm}

\section{Rank revealing of SVD-based models}

Here we demonstrate the rank revealing property of the SVD-based models. Recall that the low-rank matrix $\mat{X}$ is parametrized in the SVD format
\begin{equation*}
  \mat{X} = \mat{U}\mat{S}\mat{V}^T, \quad \mat{U}^T\mat{U} = \mat{I}_r, \quad \mat{V}\mat{V}^T = \mat{I}_{r}.
\end{equation*}
The nuclear norm of $\mat{X}$ is given as the summation of its singular values, i.e., 
\[
  \lVert \mat{X} \rVert _* = \text{trace}( \mat{X}^T \mat{X} ) = \sum_{i = 1}^r \mat{S}_{ii}.
\]
Since the nuclear norm $\lVert \mat{X} \rVert _*$ is a convex envelope of the rank of a matrix, $\text{rank}\mat{X})$, it often used as a regularization term in optimization problems to estimate low-rank matrices. See \cite{candes2010power,hastie2015matrix,mazumder2010spectral} and references therein for further discussions. 
Drawing an analogy with the nuclear norm regularization, a natural choice of the prior distribution for the singular values of the SVD-based model is to use the exponential of the negative of the nuclear norm of $\mat{X}$, which leads to 
\[
  P(S) \propto \exp\left( - \lambda \lVert \mat{X} \rVert _*  \right) = \exp\left( - \lambda \sum_{i = 1}^r \mat{S}_{ii} \right),
\]
where $\lambda > 0$ is a penalty parameter. 

\paragraph{Connection with nuclear norm regularization} Given a likelihood function $P(y | \mat{U}, \mat{S}, \mat{V})$,  the {\it maximum a posteriori} (MAP) estimate can be expressed As
\[
\argmax_{\mat{U}, \mat{S}, \mat{V}} P( \mat{U}, \mat{S}, \mat{V} | y ) =   \argmax_{\mat{U}, \mat{S}, \mat{V}} P( y | \mat{U}, \mat{S}, \mat{V} ) \pi_{\mathcal{H}}(\mat{U})P(\mat{S}) \pi_{\mathcal{H}}(\mat{V}),
\]
where $\pi_{\mathcal{H}}(\mat{U})$ and $\pi_{\mathcal{H}}(\mat{V})$ are uniform in the Hausdorff measure and implicitly imply the constraints $\mat{U}^T\mat{U} = \mat{I}_r$ and $\mat{V}\mat{V}^T = \mat{I}_{r}$ on Stiefel manifolds. In this way, the MAP estimate can be equivalently expressed as
\[
  \argmin_{\mat{U}, \mat{S}, \mat{V}}  - \log P( y | \mat{U}, \mat{S}, \mat{V} ) + \lambda \sum_{i = 1}^r \mat{S}_{ii}, \quad \text{subject\;to} \quad \mat{U}^T\mat{U} = \mat{I}_r, \quad \mat{V}\mat{V}^T = \mat{I}_{r}.
\]
The above formulation explains that the MAP estimate of our proposed SVD-based model is equivalent to a matrix estimation problem regularized by the nuclear norm, which is a widely-accepted way to impose low-rank properties. 

\paragraph{Numerical demonstration}  The above justification hinted that the SVD-based model together with the nuclear-norm-type prior can be capable of determining the effective rank of a Bayesian matrix estimation problem. To demonstrate this rank revealing property, we use the Case \#1 of the synthetic example presented in Section 3 of the paper. Here we construct a true matrix \( \mat{X}_\text{true} = \mat{A} \mat{B}^T \), where entries of $\mat{A} \in \mathbb{R}^{100 \times 10}, \mat{B} \in \mathbb{R}^{60 \times 10}$ are drawn from i.i.d. standard Gaussian, and partially observed data $y_{ij} = \mat{X}_{ij}$ are given by randomly selecting a subset of $40\%$ of matrix indices. 

\begin{figure}[h]
  \centering
  \vspace{-6pt}
  \includegraphics[height=0.2\textwidth, trim = {3em, 0, 3em, 0},  clip]{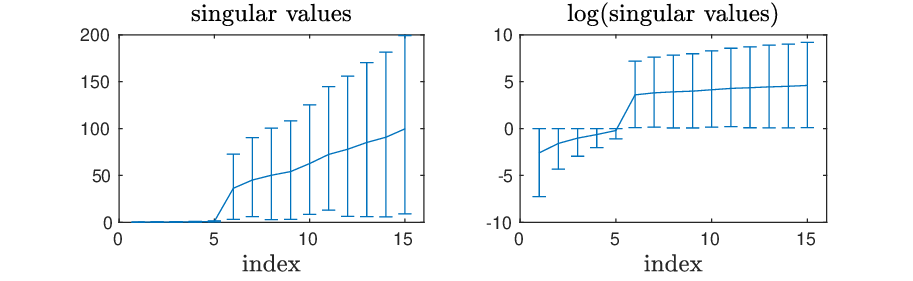}
  \caption{Demonstration of the rank revealing property using Case \#1 of the synthetic example, with $40\%$ data sampling rate. Left: quantiles of the posterior singular values. Right: quantiles of the posterior singular values on the natural logarithmic scale. The solid lines represent the median and the error bars represent $10\%$ and $90\%$ quantiles.}\label{fig:svd_case1}
\end{figure}

The true matrix is rank $10$ by construction, we impose a maximum rank $15$ in the parameterization and run the proposed geodesic HMC algorithm to sample the posterior distribution. The $10\%$ quantile, median and $90\%$ quantile of the posterior singular values are reported in Figure \ref{fig:svd_case1}. Here, we observe that beyond rank 10, the estimated singular values are close to zero and the sizes of the corresponding credible intervals are near zero. This is a clear indication that the actual rank of the estimated matrix is about $10$.
Using our proposed SVD-based models, the columns of $\mat{U}$ and $\mat{V}$ that do not contribute to the estimation problem will be automatically assigned near zero singular values in this example. 

\section{Sampling efficiency}

In this section, we provide a detailed summary of the sampling efficiency of the geodesic HMC and Gibbs sampling respectively applied to SVD-based models and the $\mat{A}\mat{B}^T$ model. To preserve anonymity, the git repositories containing the code for reproducing the results will be made public after the completion of the review process. URLs and instructions for running the code will also be provided in the final submission.

Table \ref{tab:summary} summarizes the CPU time per iteration. Here the number of steps $T$ and the time step $\epsilon$ used by geodesic HMC are automatically tuned by the No-U-Turn-Sampler variant of HMC. Each iteration of Gibbs consists of a full sweep of updates of $(\mat{A}, \mat{B}, \gamma, \tau)$. The CPU times are measured on a DELL workstation with dual Intel Xeon(R) E5-2680 v4 CPU.

\begin{table}[h]
\caption{Summary of the CPU time per iteration of the geodesic HMC and the Gibbs sampling, measured over 30 repetitions. The model $\mat{A}\mat{B}^T$ is sampled by the Gibbs sampling. The SVD-based models are sampled by geodesic HMC. For table entries shown with ``$\nicefrac{}{}$,'' the left and the right represent results for the $10\%$ and $40\%$ data sampling rates, respectively.}
\label{tab:summary}
\footnotesize
\begin{center}
\renewcommand{\arraystretch}{1.3}
\hspace{-5em}
\begin{tabular}{l|c|c|c}
\toprule
&  \multicolumn{3}{c}{mean$\pm$standard deviation of CPU time (in $10^{-3}$ second)}\\
\cline{2-4}
& $\mat{A}\mat{B}^T$& SVD & $\ast$-SVD \\
\hline
 Case \#1  & $\nicefrac{4.1\pm1}{5.6\pm1}$ & $\nicefrac{19\pm 2.1 }{26 \pm 2.9}$ & - \\
\cline{2-4}
 Case \#2  & $\nicefrac{4.1\pm0.3}{4.9\pm0.8}$ & $\nicefrac{67\pm 14 }{189 \pm 12}$ & $\nicefrac{85 \pm 56 }{159 \pm 16}$\\
\cline{2-4}
 Case \#3  & $\nicefrac{4.0\pm0.3}{5.2\pm0.6}$ & $\nicefrac{36\pm0.7}{158\pm2.6}$ & $\nicefrac{0.7\pm0.1}{0.7\pm0.2}$ \\
\hline
 Mice  & $\nicefrac{47\pm3.1}{65\pm3.6}$ &  $\nicefrac{650 \pm 12}{940\pm350}$ & $\nicefrac{670 \pm 43}{878\pm305}$ \\
\hline
MovieLens & \scriptsize $1400 \pm 340$ & \scriptsize $2100 \pm 220 $ & \scriptsize $2400 \pm 180 $  \\
\bottomrule
\end{tabular}
\end{center}
\end{table}

\begin{figure}[h]
  \centering
  \vspace{-6pt}
  \includegraphics[width=0.9\textwidth, trim = {4em, 0, 2em, 0},  clip]{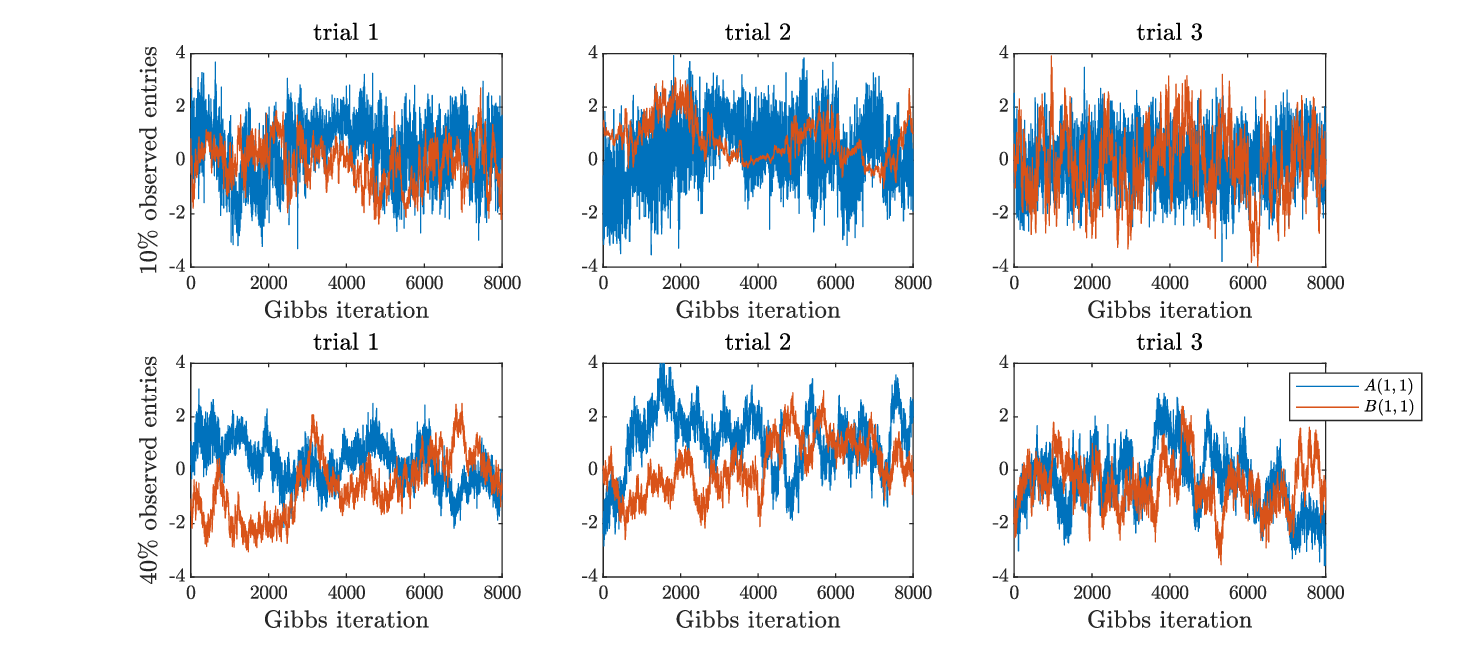}
  \caption{Three random experiments of Case \#1 of the synthetic example. MCMC traces generated by Gibbs using the $\mat{A}\mat{B}^T$ model. Top row: $10\%$ data sampling rate. Bottom row: $40\%$ data sampling rate. }\label{fig:trace_case1_gibbs}
\end{figure}

\begin{figure}[h]
  \centering
  \vspace{-6pt}
  \begin{subfigure}[b]{0.95\linewidth}
  \includegraphics[width=0.9\textwidth, trim = {6em, 0, 4em, 0},  clip]{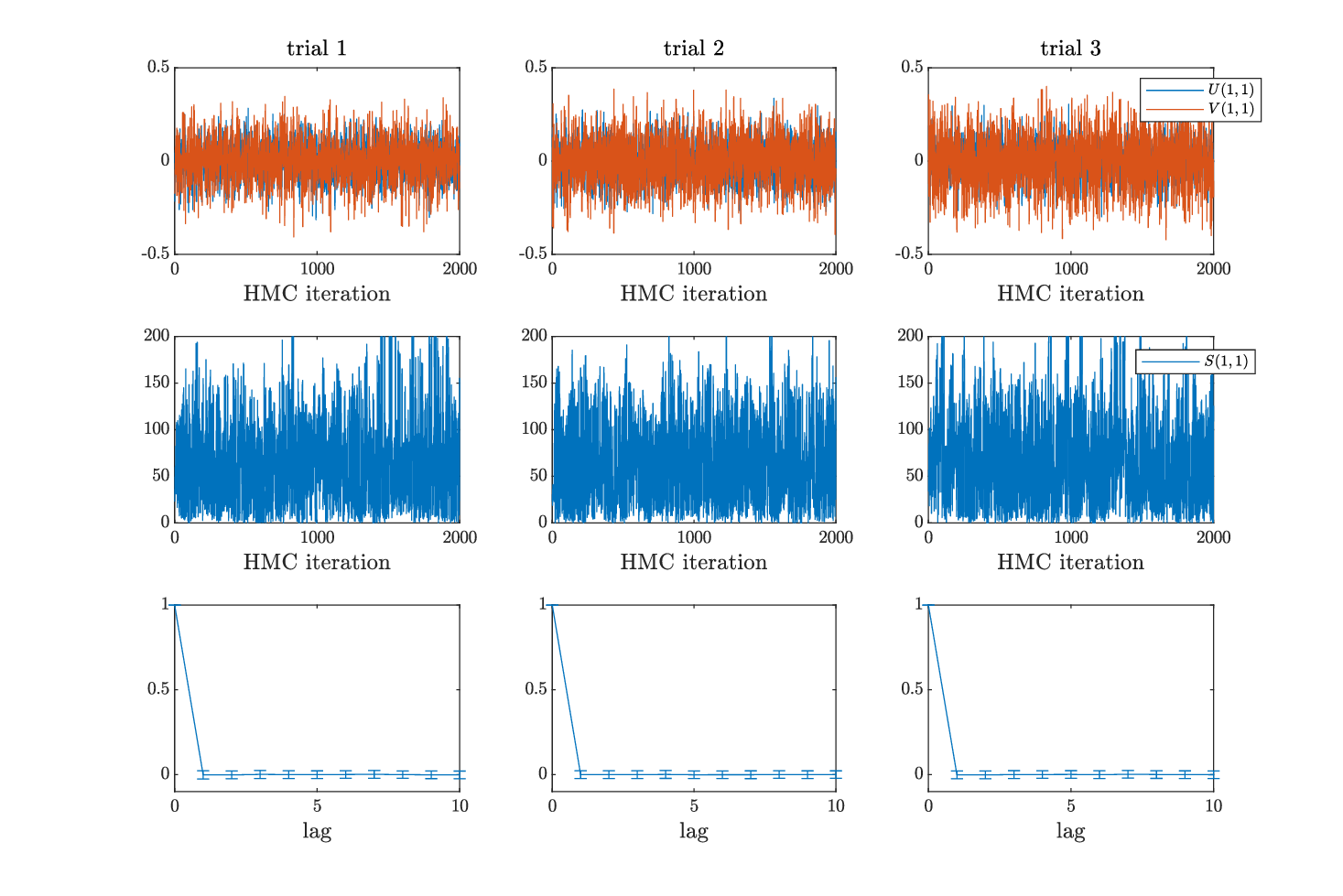}
  \caption{$10\%$ data sampling rate}
  \end{subfigure}

  \begin{subfigure}[b]{0.95\linewidth}
  \includegraphics[width=0.9\textwidth, trim = {6em, 0, 4em, 0},  clip]{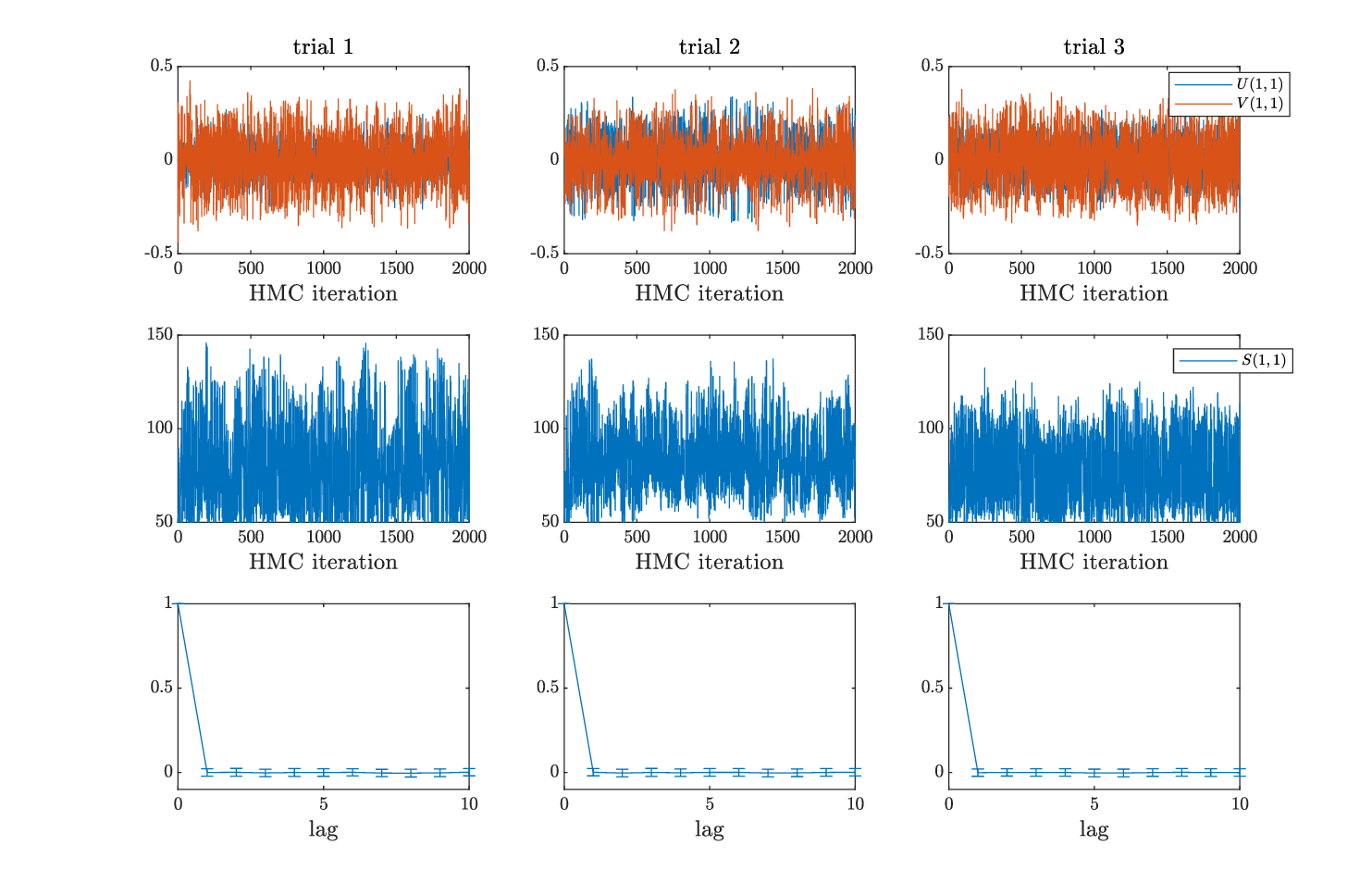}
  \caption{$40\%$ data sampling rate}
  \end{subfigure}
  \caption{Three random experiments of Case \#1 of the synthetic example. The SVD model and the geodesic HMC are used here.  Top and middle row: MCMC traces. Bottom row: average autocorrelation function and standard derivation estimated over all parameters.}\label{fig:trace_case1_hmc}
\end{figure}

\begin{figure}[h]
  \centering
  \vspace{-6pt}
  \begin{subfigure}[b]{0.95\linewidth}
    \includegraphics[width=0.9\textwidth, trim = {6em, 0, 4em, 0},  clip]{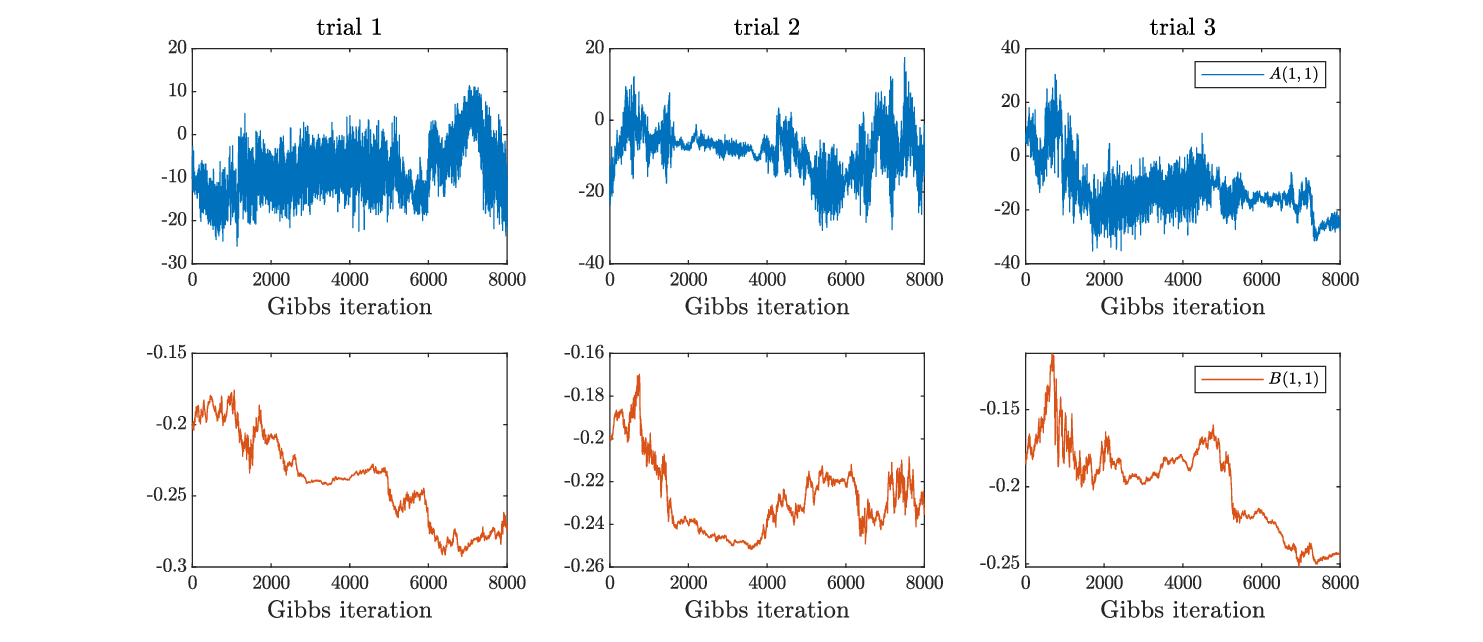}
  \caption{$10\%$ data sampling rate}
  \end{subfigure}

  \begin{subfigure}[b]{0.95\linewidth}
  \includegraphics[width=0.9\textwidth, trim = {6em, 0, 4em, 0},  clip]{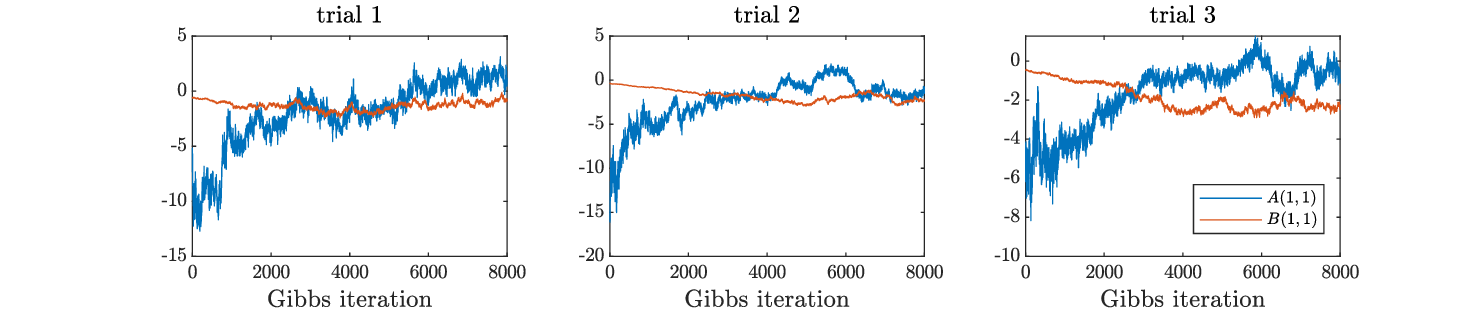}
  \caption{$40\%$ data sampling rate}
  \end{subfigure}
  \caption{Three random experiments of Case \#2 of the synthetic example. MCMC traces generated by Gibbs using the $\mat{A}\mat{B}^T$ model. }\label{fig:trace_case2_gibbs}
\end{figure}

\begin{figure}[h]
  \centering
  \vspace{-6pt}
  \begin{subfigure}[b]{0.95\linewidth}
  \includegraphics[width=0.9\textwidth, trim = {6em, 0, 4em, 0},  clip]{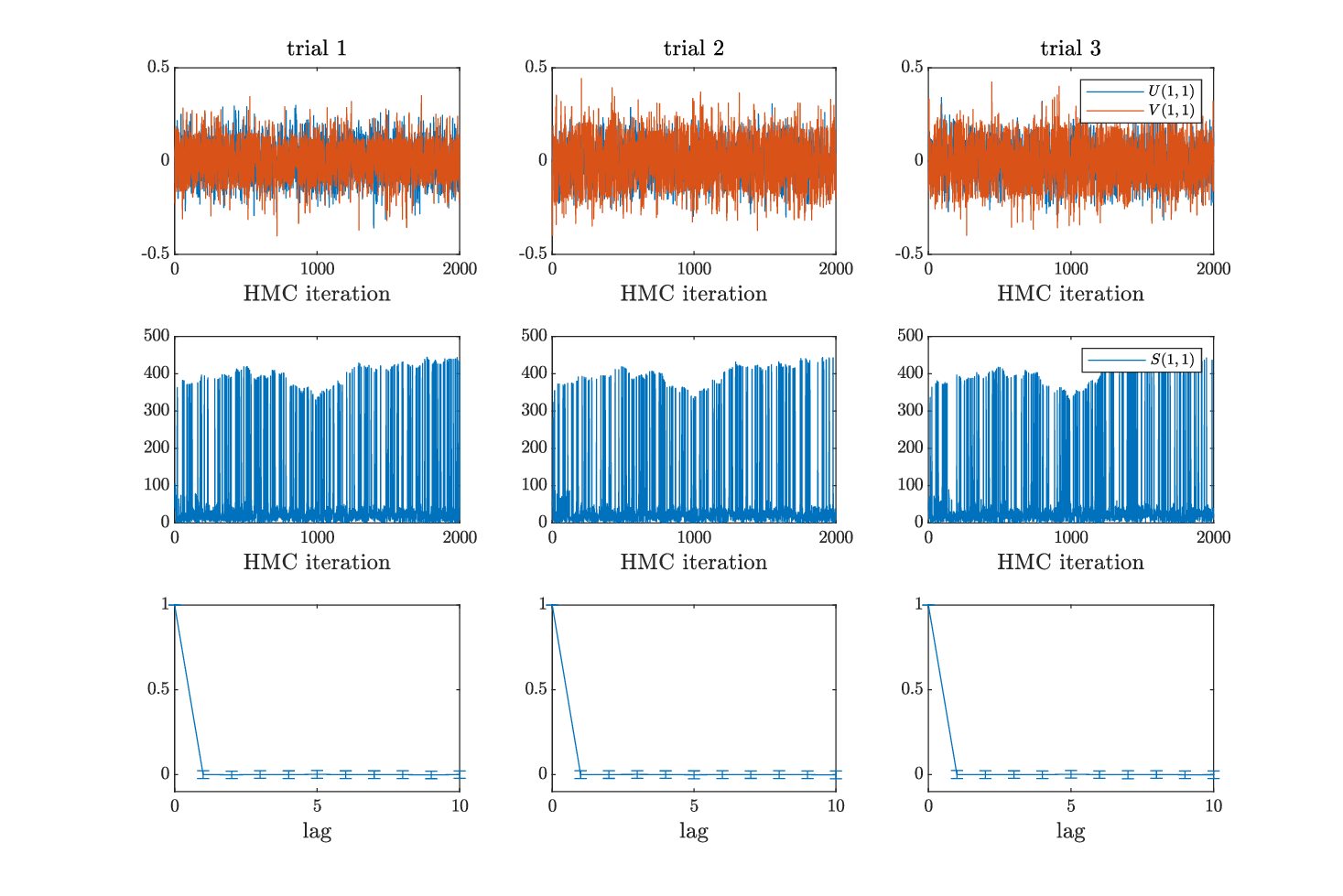}
  \caption{$10\%$ data sampling rate}
  \end{subfigure}

  \begin{subfigure}[b]{0.95\linewidth}
  \includegraphics[width=0.9\textwidth, trim = {6em, 0, 4em, 0},  clip]{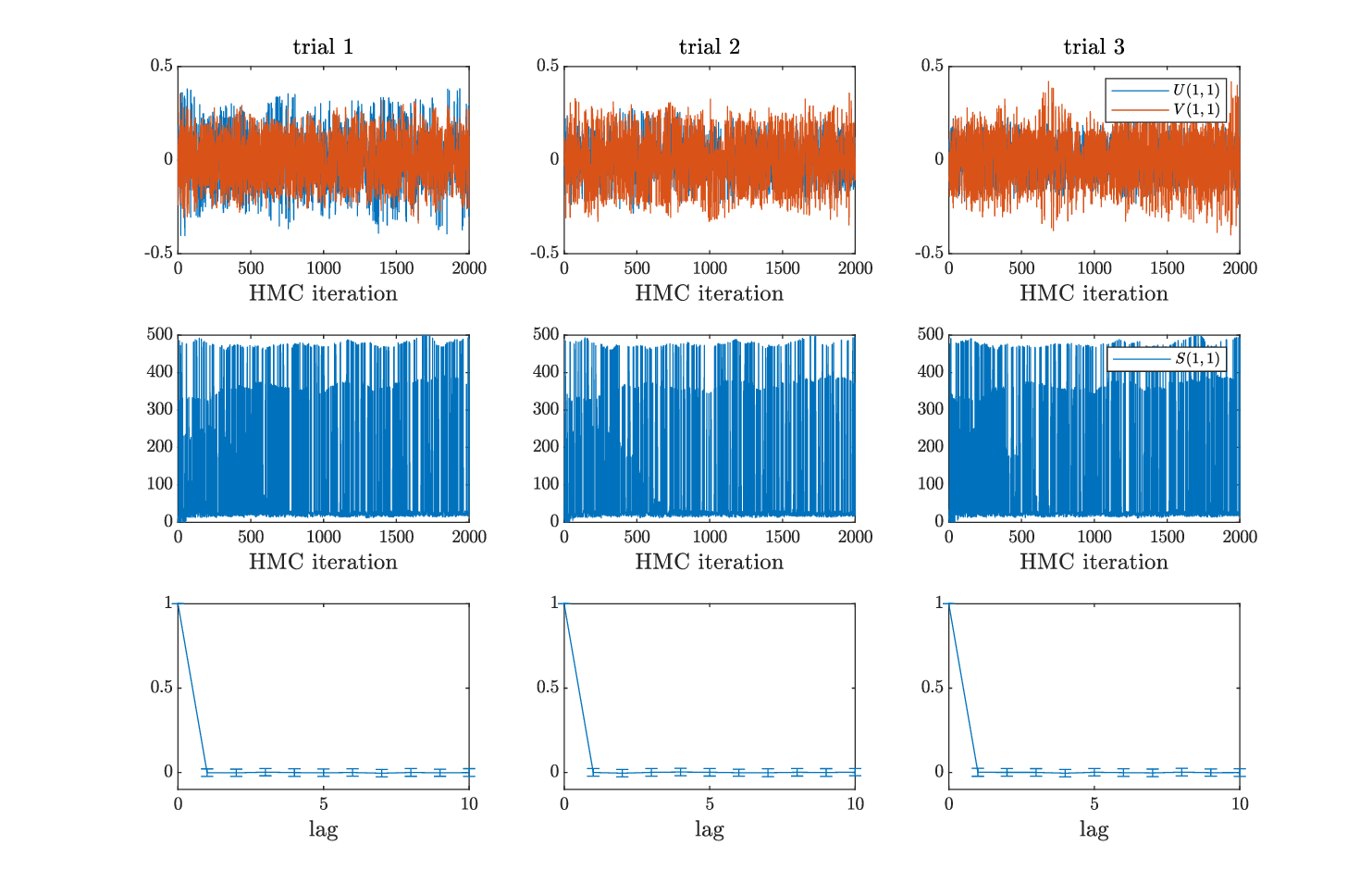}
  \caption{$40\%$ data sampling rate}
  \end{subfigure}
  \caption{Three random experiments of Case \#2 of the synthetic example. The S-SVD model and the geodesic HMC are used here. Top and middle row: MCMC traces. Bottom row: average autocorrelation function and standard derivation estimated over all parameters.}\label{fig:trace_case2_s_hmc}
\end{figure}

\begin{figure}[h]
  \centering
  \vspace{-6pt}
  \begin{subfigure}[b]{0.95\linewidth}
  \includegraphics[width=0.9\textwidth, trim = {6em, 0, 4em, 0},  clip]{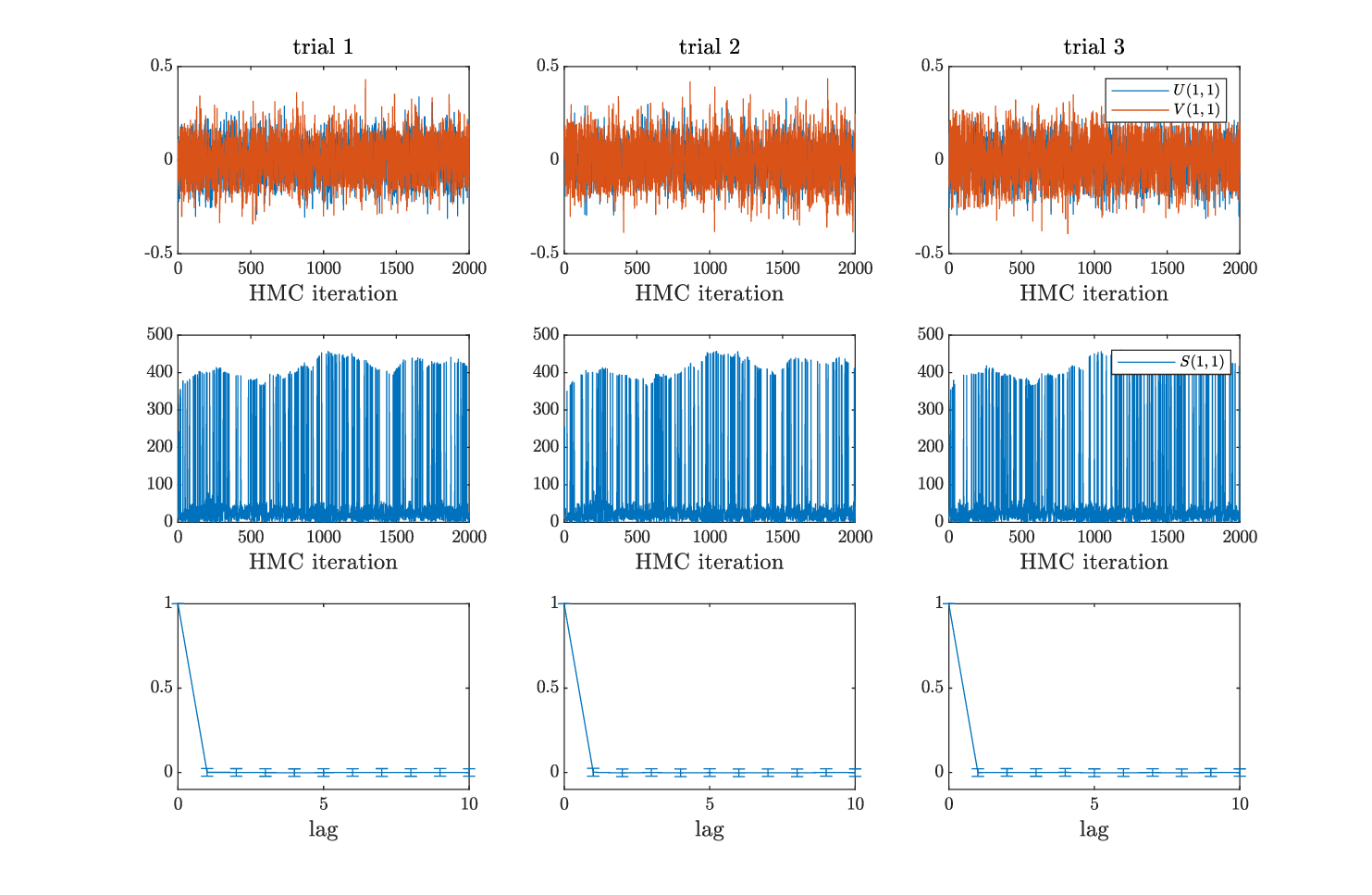}
  \caption{$10\%$ data sampling rate}
  \end{subfigure}

  \begin{subfigure}[b]{1\linewidth}
  \includegraphics[width=0.9\textwidth, trim = {6em, 0, 4em, 0},  clip]{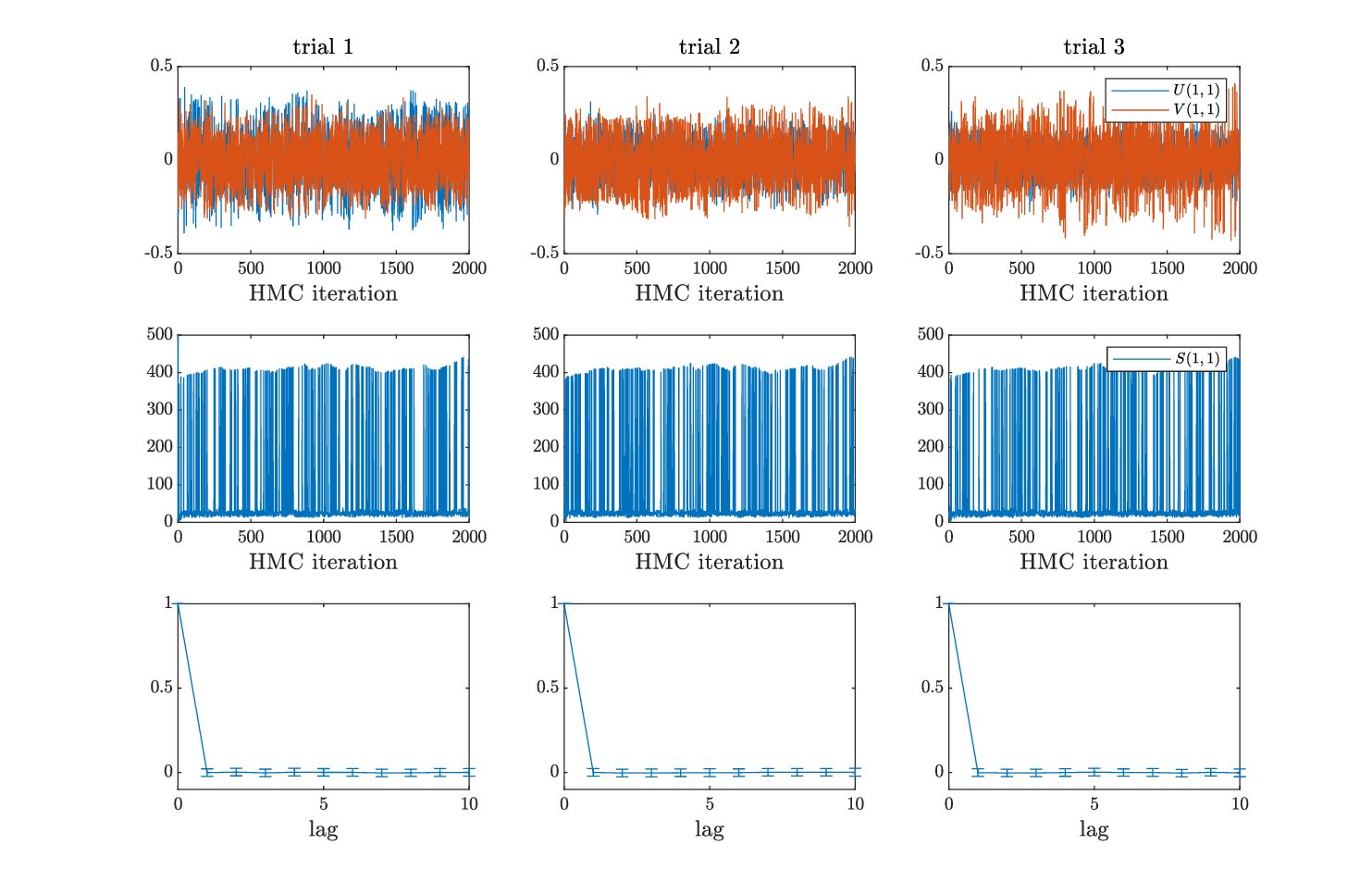}
  \caption{$40\%$ data sampling rate}
  \end{subfigure}
  \caption{Three random experiments of Case \#2 of the synthetic example. The SVD model and the geodesic HMC are used here. Top and middle row: MCMC traces. Bottom row: average autocorrelation function and standard derivation estimated over all parameters.}\label{fig:trace_case2_g_hmc}
\end{figure}

\begin{figure}[h]
  \centering
  \vspace{-6pt}
  \begin{subfigure}[b]{0.95\linewidth}
  \includegraphics[width=0.9\textwidth, trim = {6em, 0, 4em, 0},  clip]{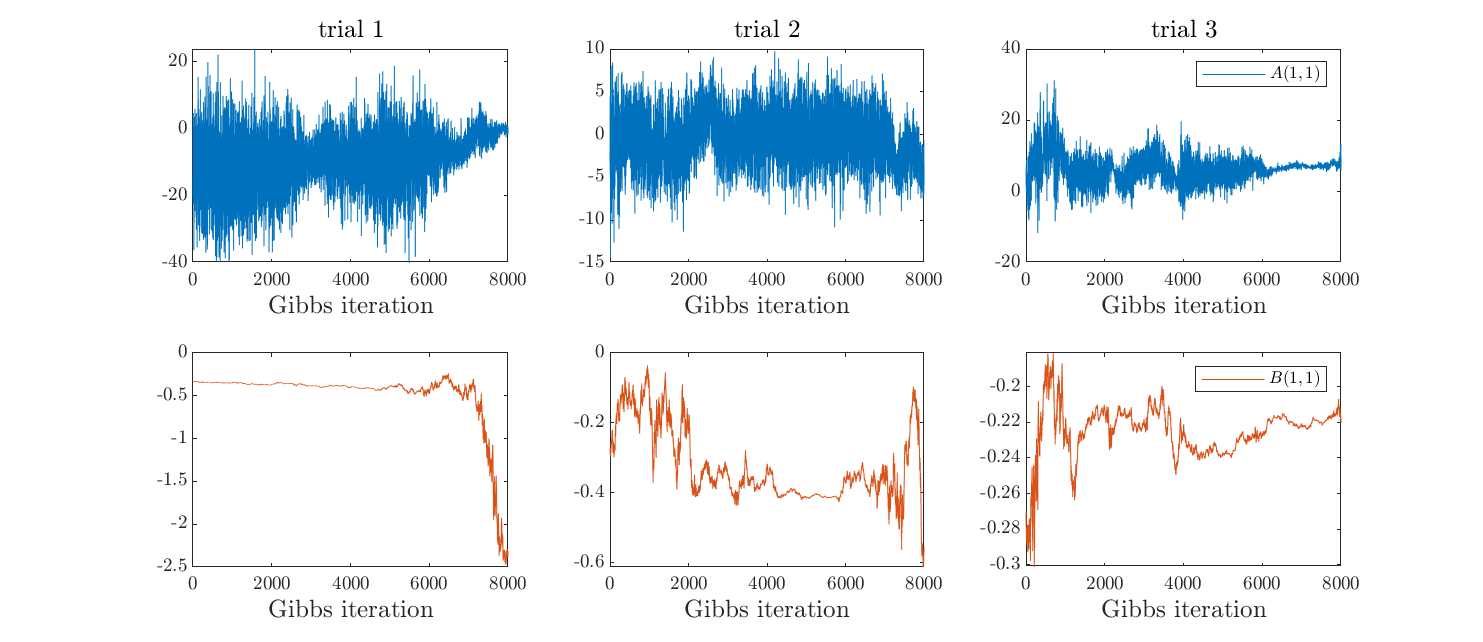}
  \caption{$10\%$ data sampling rate}
  \end{subfigure}

  \begin{subfigure}[b]{0.95\linewidth}
  \includegraphics[width=0.9\textwidth, trim = {6em, 0, 4em, 0},  clip]{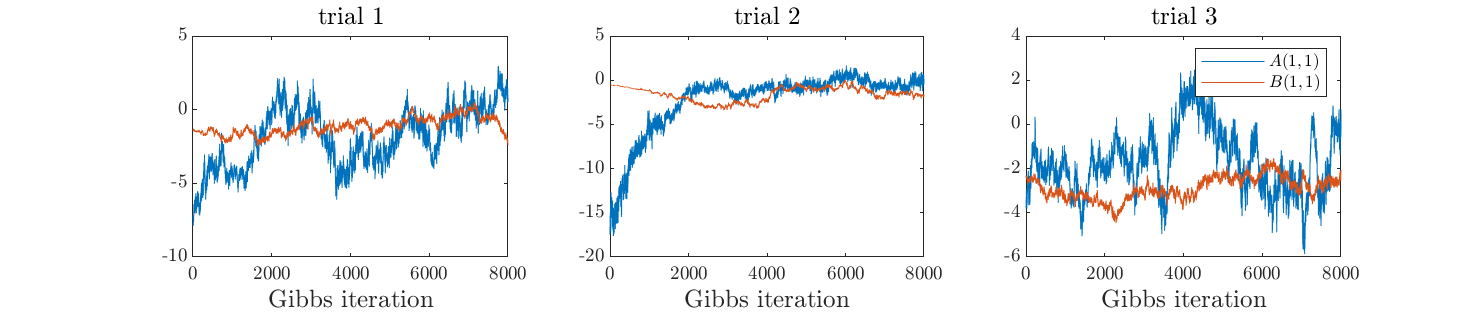}
  \caption{$40\%$ data sampling rate}
  \end{subfigure}
  \caption{Three random experiments of Case \#3 of the synthetic example. MCMC traces generated by Gibbs using the $\mat{A}\mat{B}^T$ model. }\label{fig:trace_case3_gibbs}
\end{figure}

\begin{figure}[h]
  \centering
  \vspace{-6pt}
  \begin{subfigure}[b]{0.95\linewidth}
  \includegraphics[width=0.9\textwidth, trim = {6em, 0, 4em, 0},  clip]{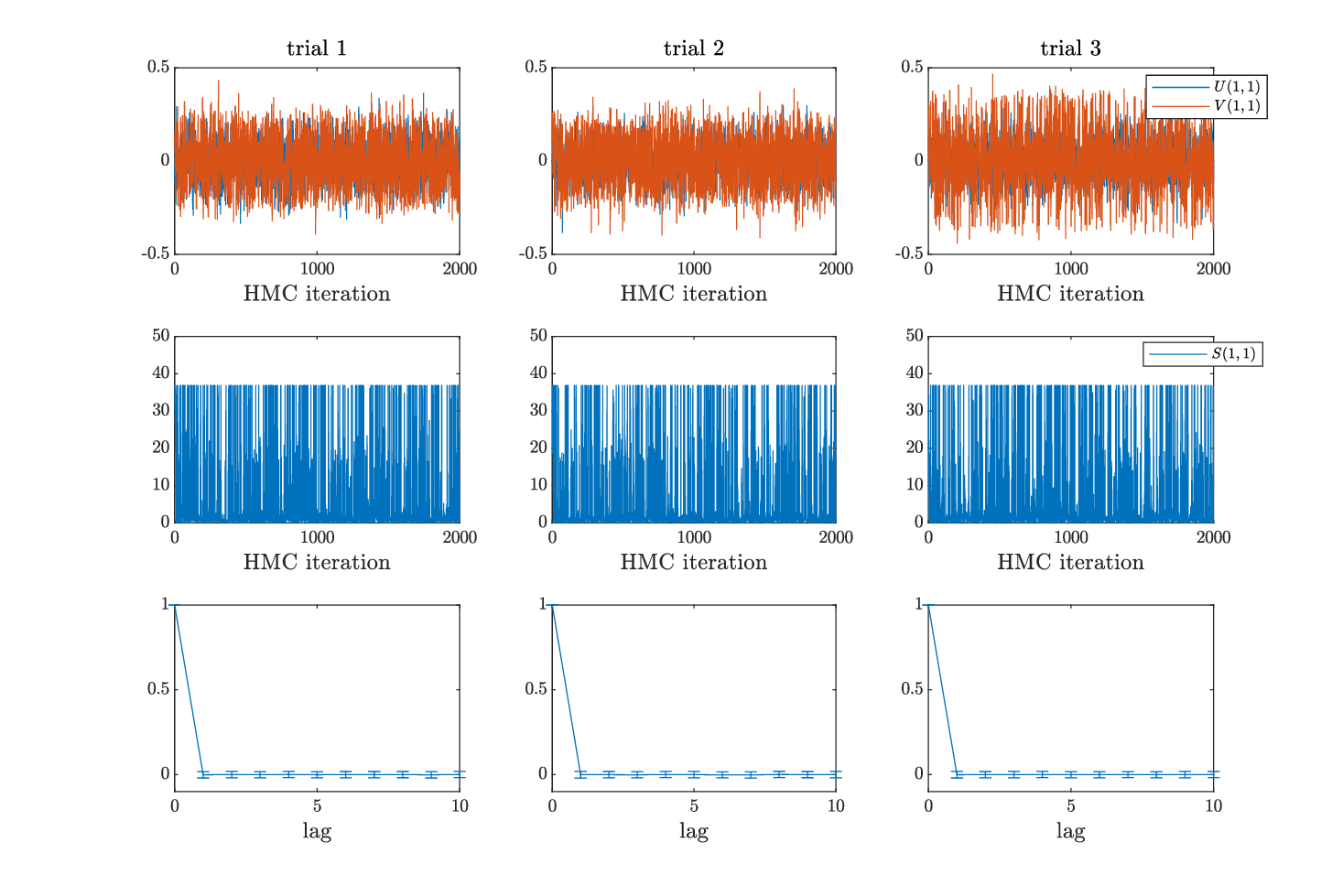}
  \caption{$10\%$ data sampling rate}
  \end{subfigure}

  \begin{subfigure}[b]{0.95\linewidth}
  \includegraphics[width=0.9\textwidth, trim = {6em, 0, 4em, 0},  clip]{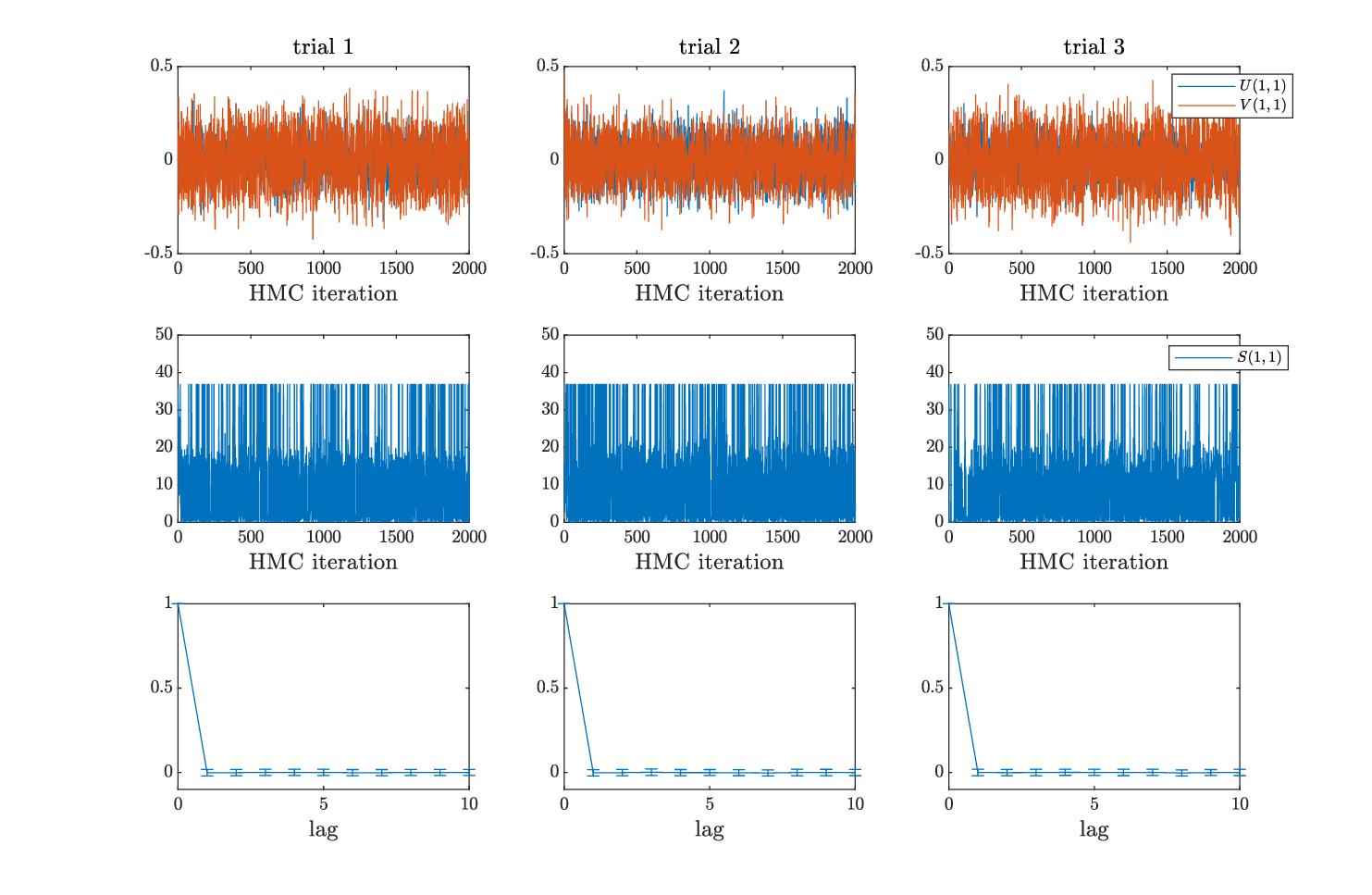}
  \caption{$40\%$ data sampling rate}
  \end{subfigure}
  \caption{Three random experiments of Case \#3 of the synthetic example. The B-SVD model and the geodesic HMC are used here. Top and middle row: MCMC traces. Bottom row: average autocorrelation function and standard derivation estimated over all parameters.}\label{fig:trace_case3_b_hmc}
\end{figure}

\begin{figure}[h]
  \centering
  \vspace{-6pt}
  \begin{subfigure}[b]{0.95\linewidth}
  \includegraphics[width=0.9\textwidth, trim = {6em, 0, 4em, 0},  clip]{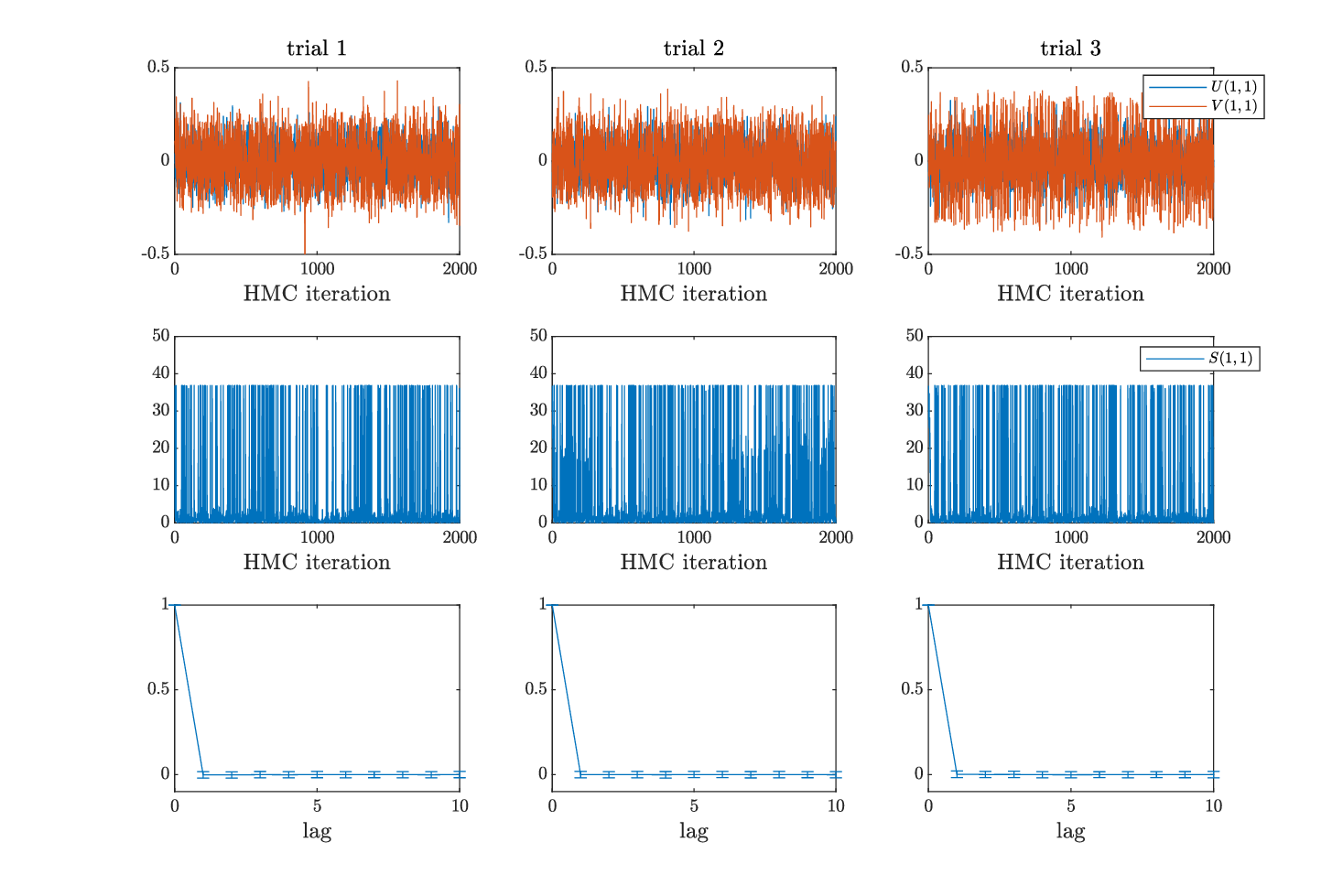}
  \caption{$10\%$ data sampling rate}
  \end{subfigure}

  \begin{subfigure}[b]{0.95\linewidth}
  \includegraphics[width=0.9\textwidth, trim = {6em, 0, 4em, 0},  clip]{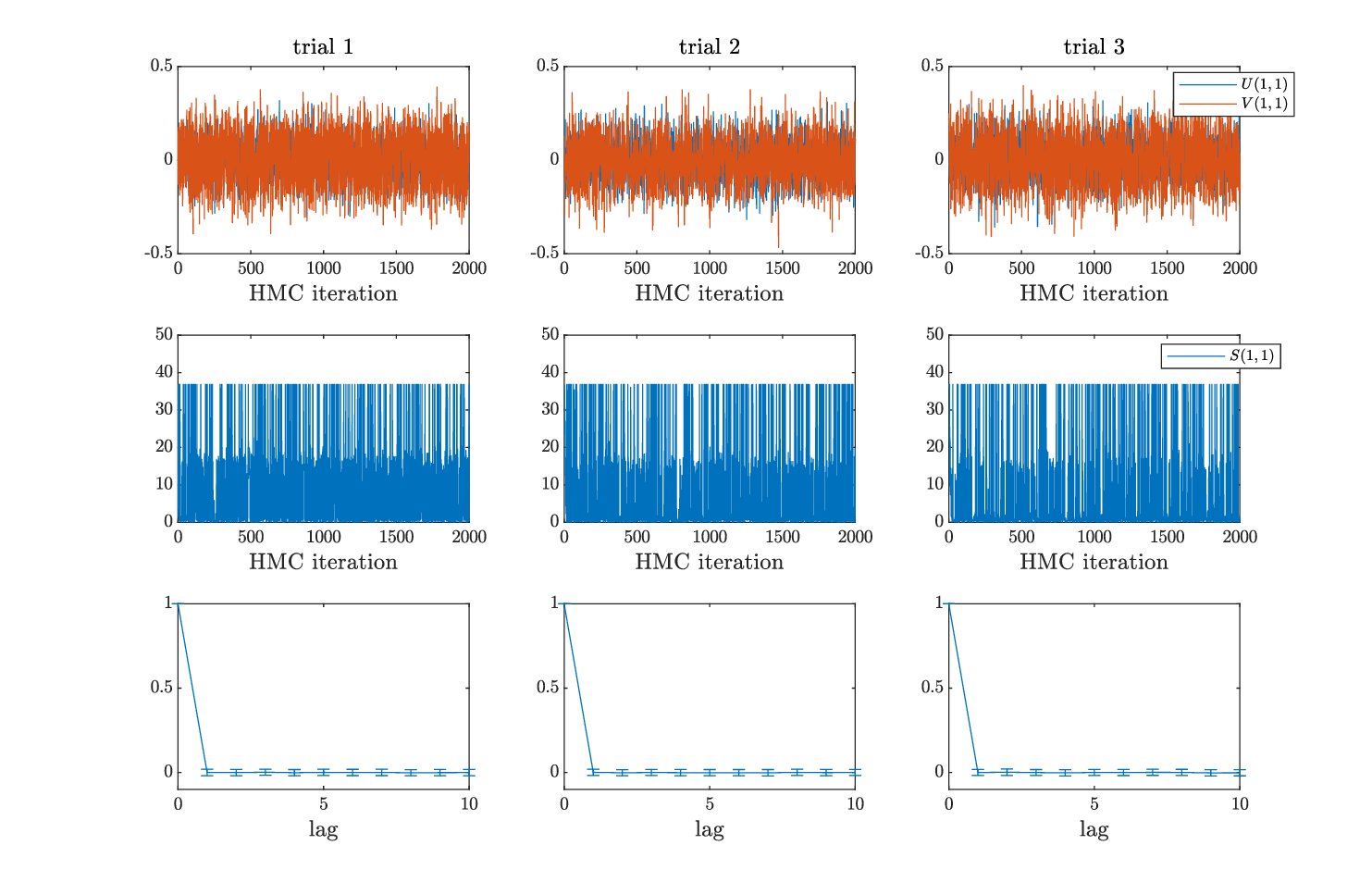}
  \caption{$40\%$ data sampling rate}
  \end{subfigure}
  \caption{Three random experiments of Case \#3 of the synthetic example. The SVD model and the geodesic HMC are used here. Top and middle row: MCMC traces. Bottom row: average autocorrelation function and standard derivation estimated over all parameters.}\label{fig:trace_case3_g_hmc}
\end{figure}

\begin{figure}[h]
  \centering
  \vspace{-6pt}
  \begin{subfigure}[b]{0.95\linewidth}
  \includegraphics[width=0.9\textwidth, trim = {6em, 0, 4em, 0},  clip]{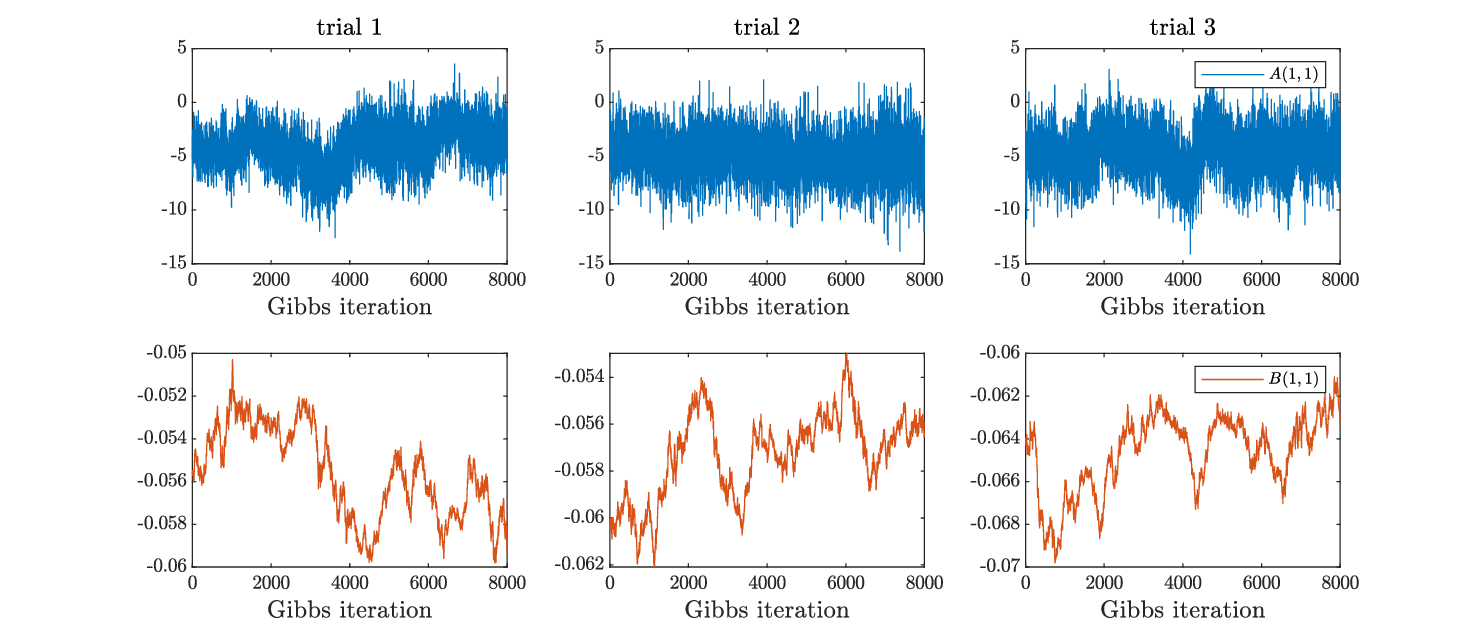}
  \caption{$10\%$ data sampling rate}
  \end{subfigure}

  \begin{subfigure}[b]{0.95\linewidth}
  \includegraphics[width=0.9\textwidth, trim = {6em, 0, 4em, 0},  clip]{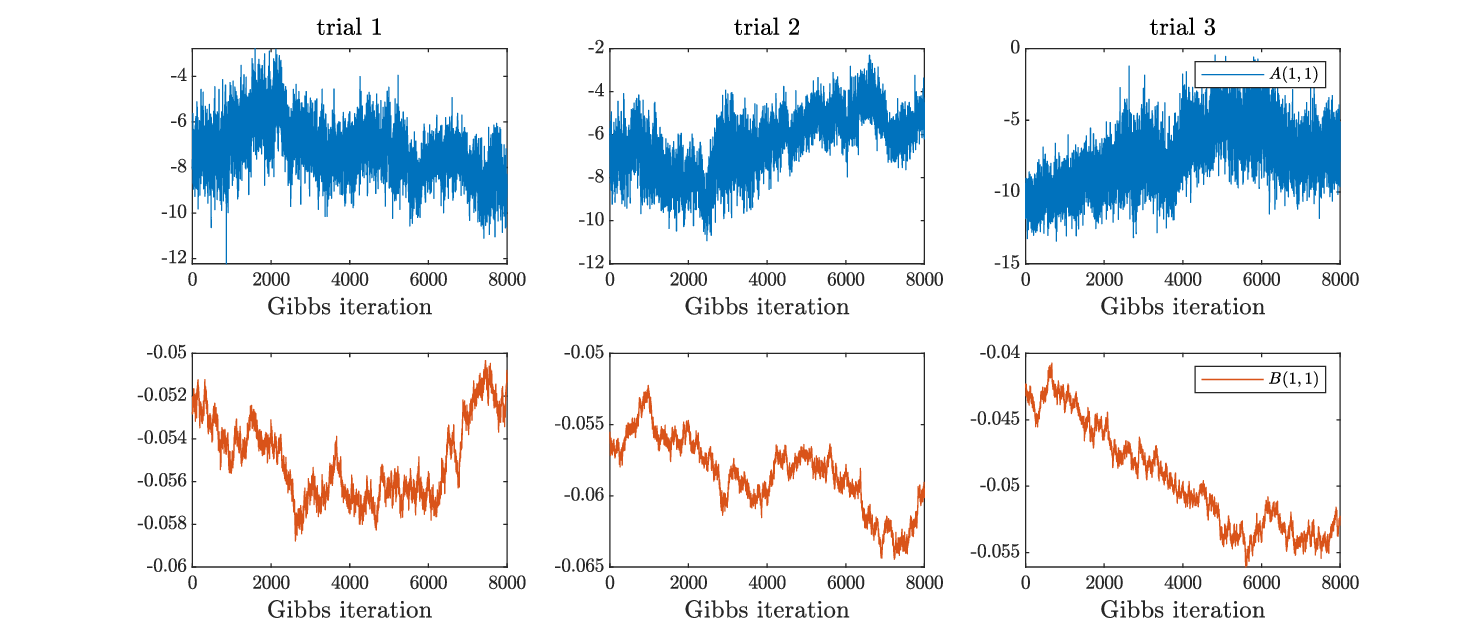}
  \caption{$40\%$ data sampling rate}
  \end{subfigure}
  \caption{Three random experiments of the mice protein expression data set. MCMC traces generated by Gibbs using the $\mat{A}\mat{B}^T$ model. }\label{fig:trace_mice_gibbs}
\end{figure}

\begin{figure}[h]
  \centering
  \vspace{-6pt}
  \begin{subfigure}[b]{0.95\linewidth}
  \includegraphics[width=0.9\textwidth, trim = {6em, 0, 4em, 0},  clip]{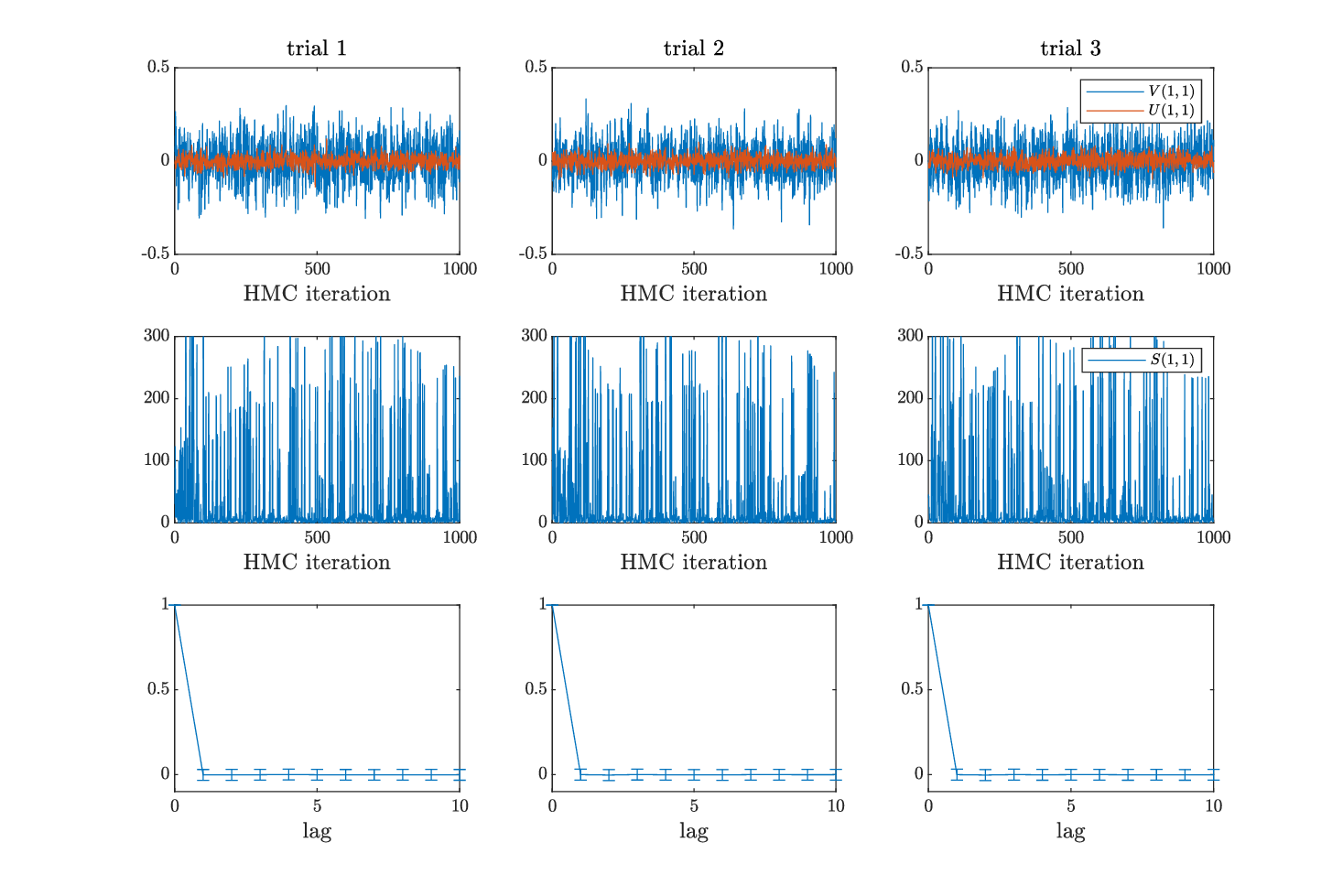}
  \caption{$10\%$ data sampling rate}
  \end{subfigure}

  \begin{subfigure}[b]{0.95\linewidth}
  \includegraphics[width=0.9\textwidth, trim = {6em, 0, 4em, 0},  clip]{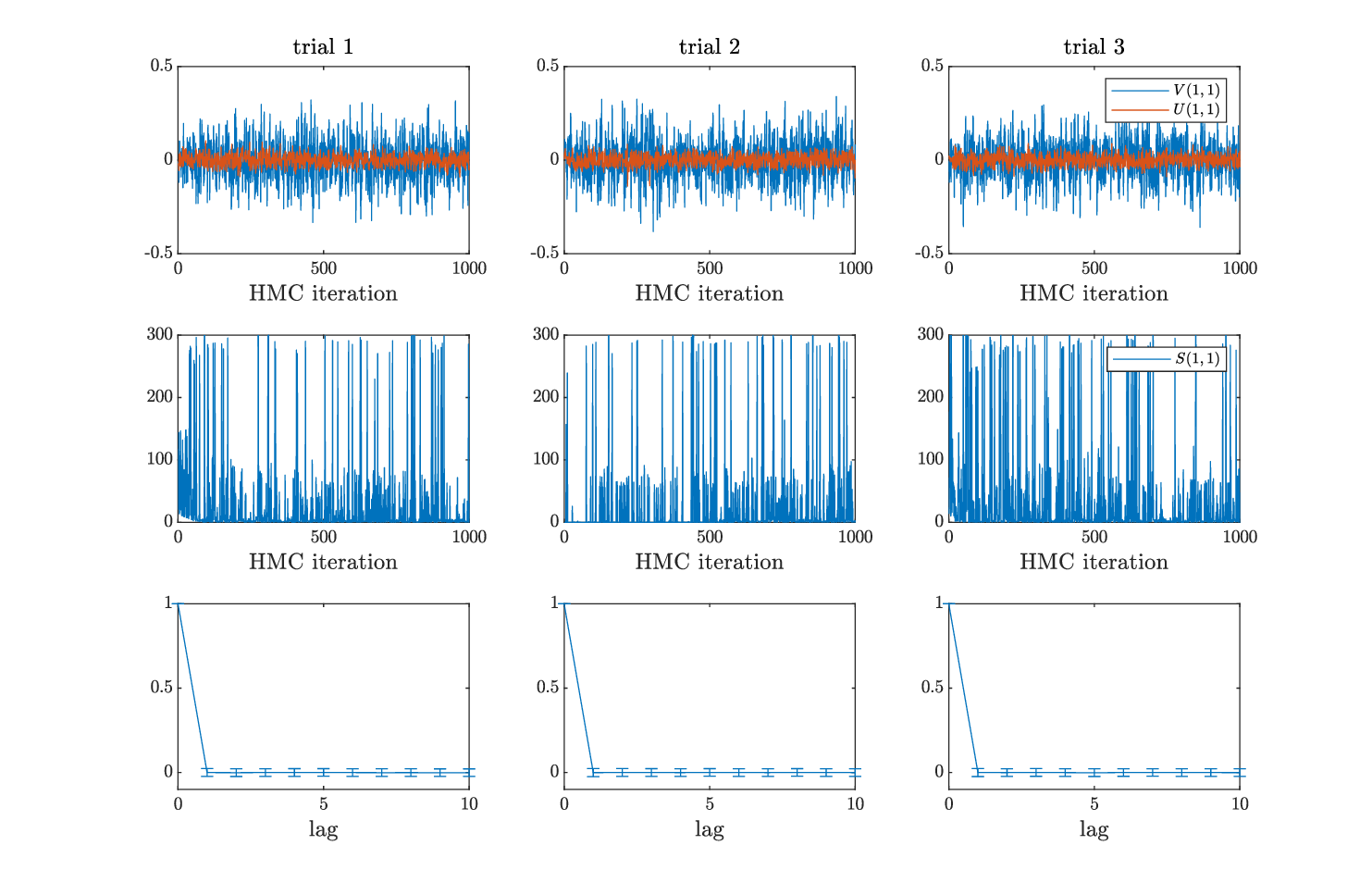}
  \caption{$40\%$ data sampling rate}
  \end{subfigure}
  \caption{Three random experiments of the mice protein expression data set. The S-SVD model and the geodesic HMC are used here. Top and middle row: MCMC traces. Bottom row: average autocorrelation function and standard derivation estimated over all parameters.}\label{fig:trace_mice_s_hmc}
\end{figure}

\begin{figure}[h]
  \centering
  \vspace{-6pt}
  \begin{subfigure}[b]{0.95\linewidth}
  \includegraphics[width=0.9\textwidth, trim = {6em, 0, 4em, 0},  clip]{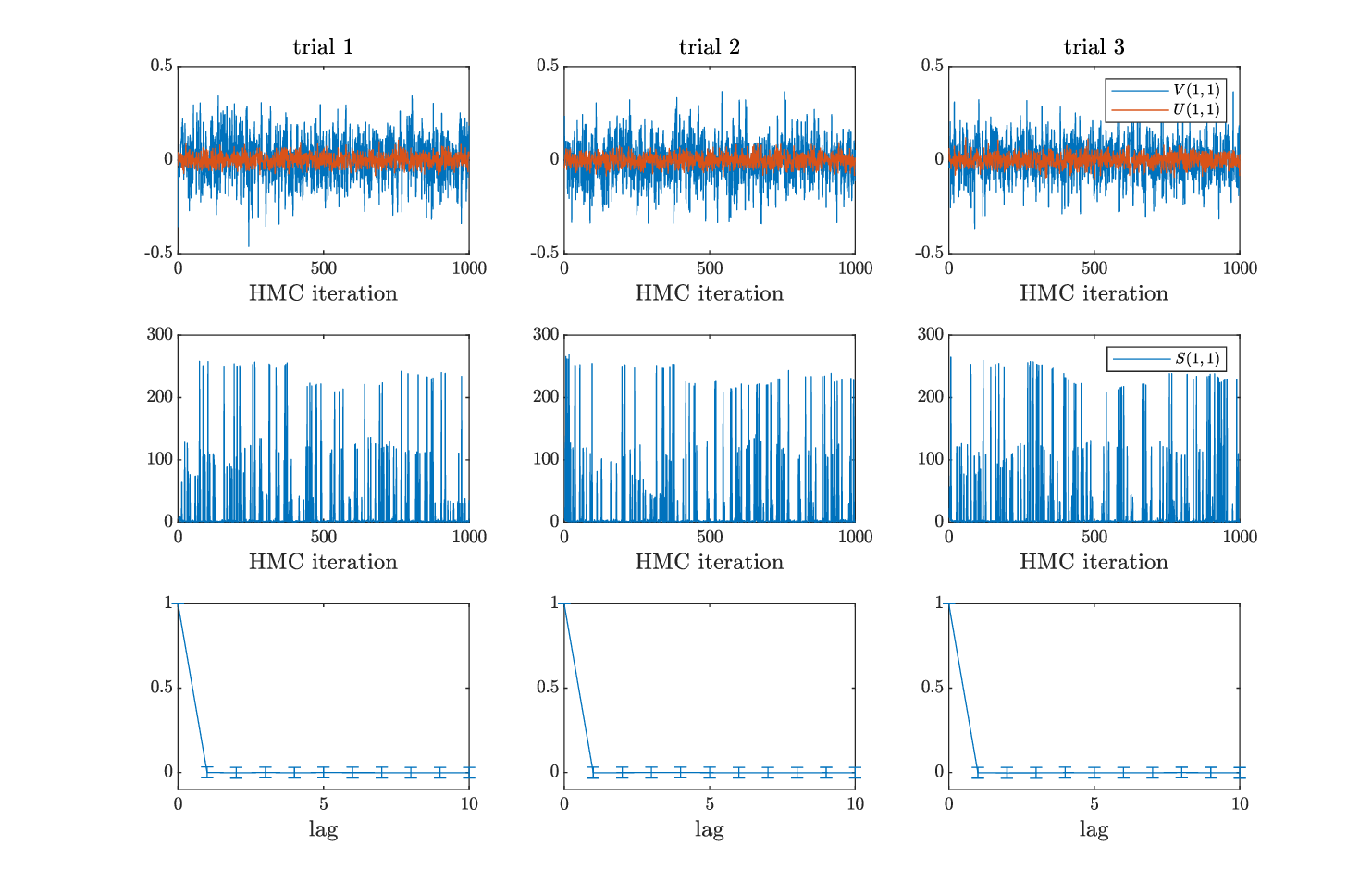}
  \caption{$10\%$ data sampling rate}
  \end{subfigure}

  \begin{subfigure}[b]{0.95\linewidth}
  \includegraphics[width=0.9\textwidth, trim = {6em, 0, 4em, 0},  clip]{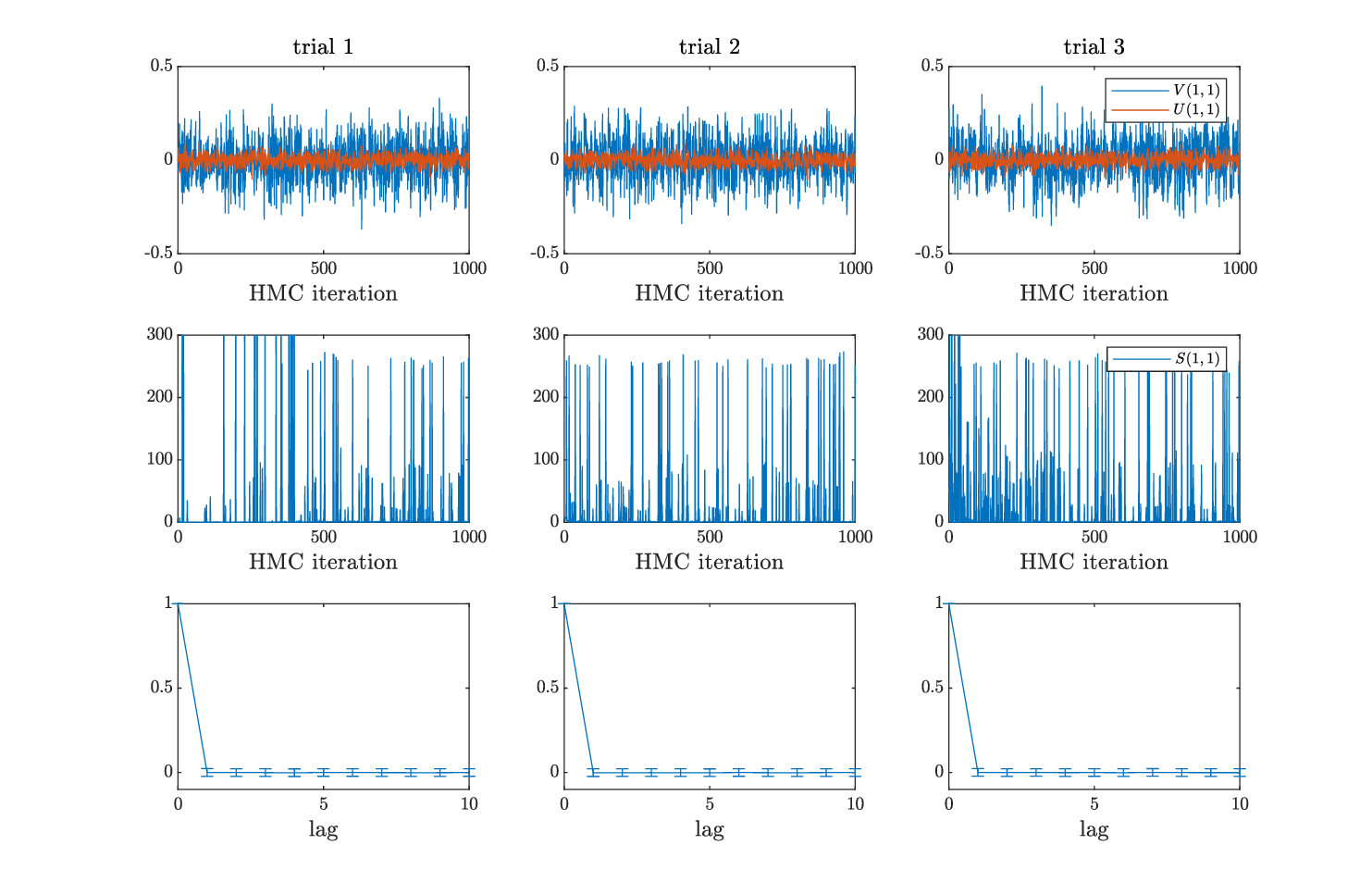}
  \caption{$40\%$ data sampling rate}
  \end{subfigure}
  \caption{Three random experiments of the mice protein expression data set. The SVD model and the geodesic HMC are used here. Top and middle row: MCMC traces. Bottom row: average autocorrelation function and standard derivation estimated over all parameters.}\label{fig:trace_mice_g_hmc}
\end{figure}

\begin{figure}[h]
  \centering
  \vspace{-6pt}
  \includegraphics[width=0.9\textwidth, trim = {6em, 0, 4em, 0},  clip]{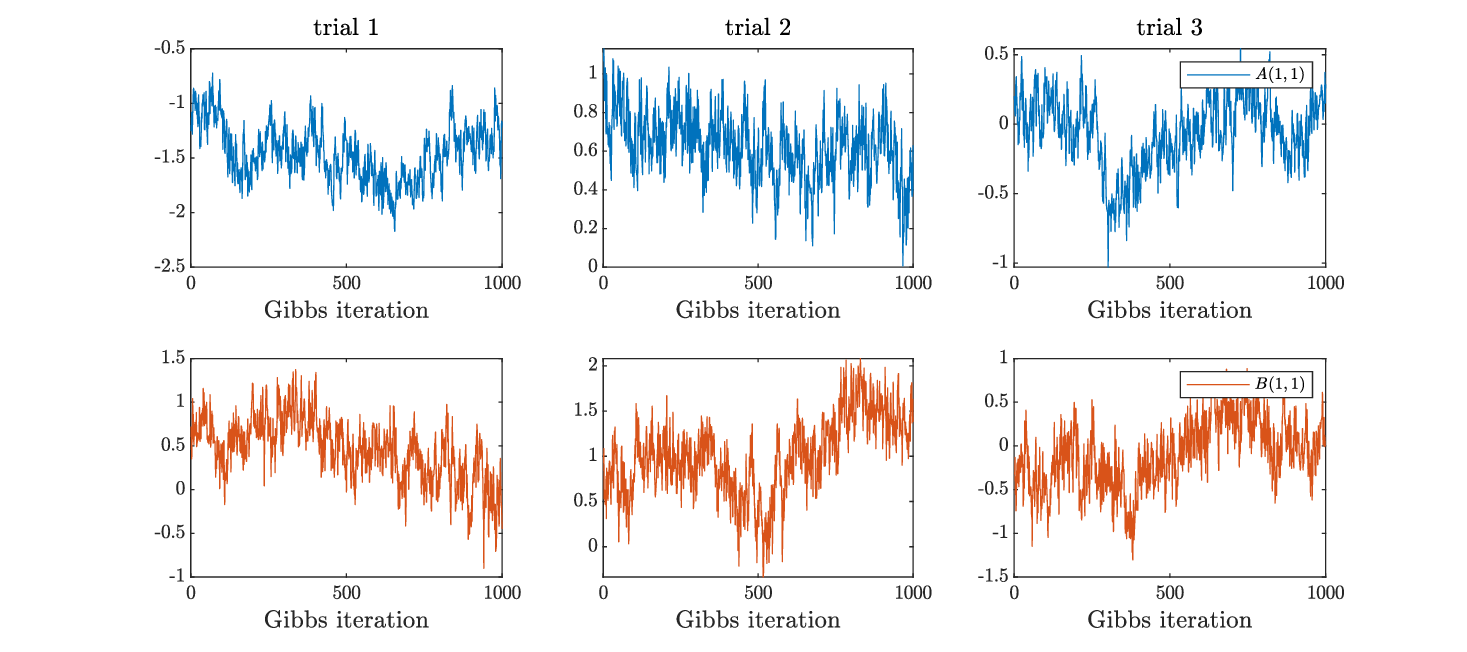}
  \caption{Three random experiments of the MovieLens data set. MCMC traces generated by Gibbs using the $\mat{A}\mat{B}^T$ model. }\label{fig:trace_movie_gibbs}
\end{figure}

\begin{figure}[h]
  \centering
  \vspace{-6pt}
  \includegraphics[width=0.9\textwidth, trim = {6em, 0, 4em, 0},  clip]{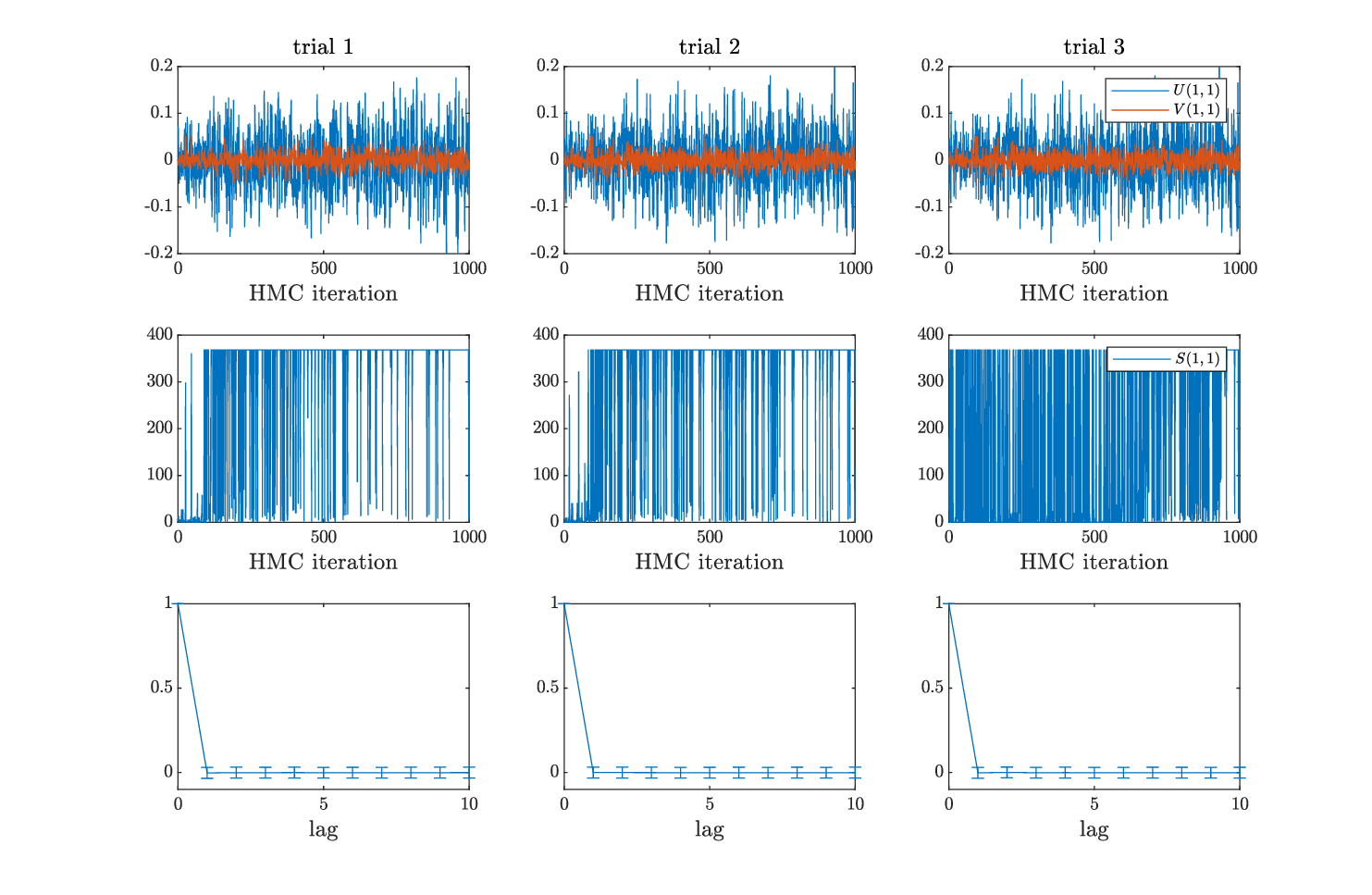}
  \caption{Three random experiments of the MovieLens data set. The S-SVD model and the geodesic HMC are used here. Top and middle row: MCMC traces. Bottom row: average autocorrelation function and standard derivation estimated over all parameters.}\label{fig:trace_movie_b_hmc}
\end{figure}

\begin{figure}[h]
  \centering
  \vspace{-6pt}
  \includegraphics[width=0.9\textwidth, trim = {6em, 0, 4em, 0},  clip]{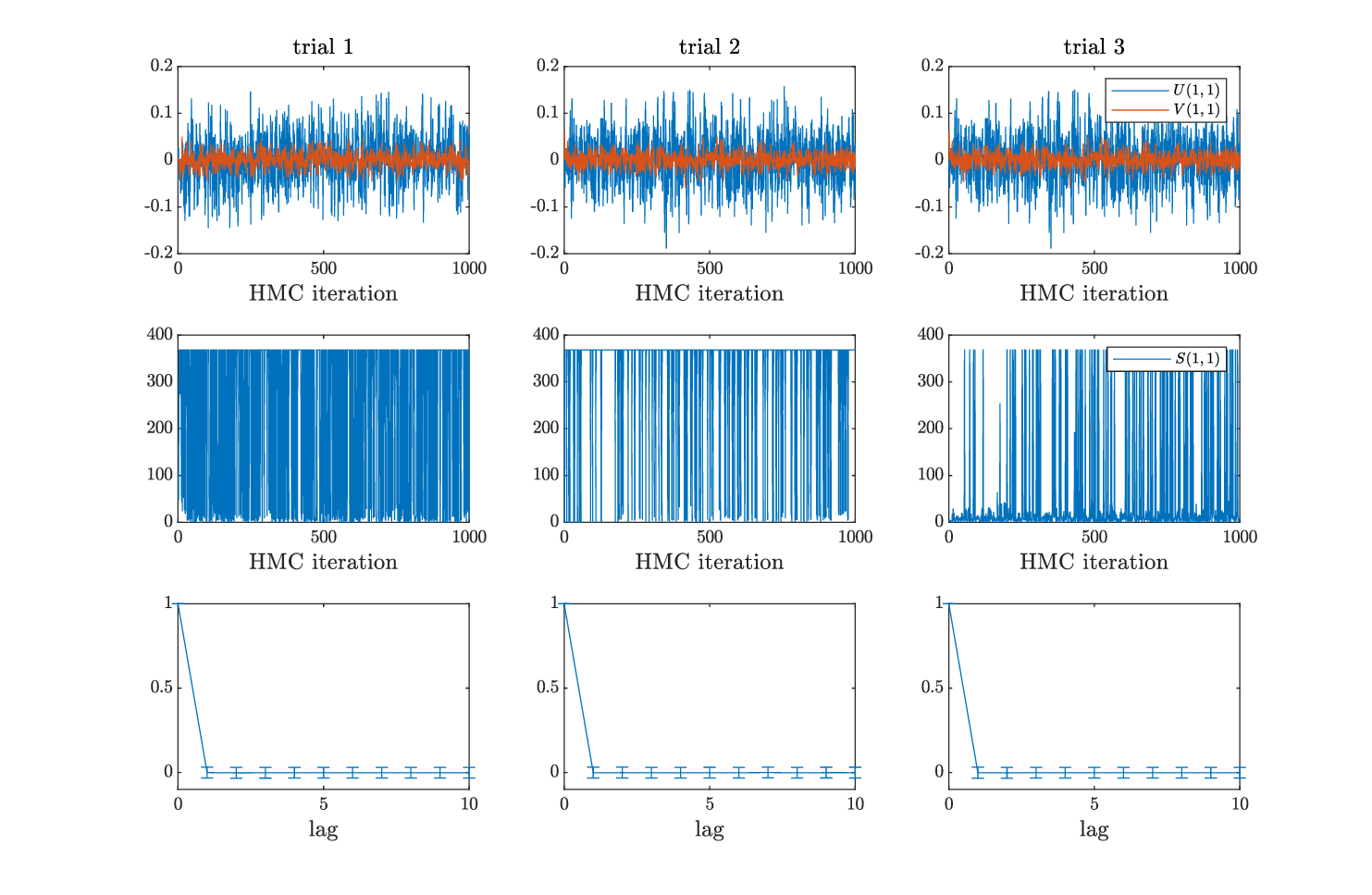}
  \caption{Three random experiments of the MovieLens data set. The SVD model and the geodesic HMC are used here. Top and middle row: MCMC traces. Bottom row: average autocorrelation function and standard derivation estimated over all parameters.}\label{fig:trace_movie_g_hmc}
\end{figure}

For each of the examples summarized in Table \ref{tab:summary}, we also show the MCMC traces generated by the corresponding samplers in Figures \ref{fig:trace_case1_gibbs}--\ref{fig:trace_movie_g_hmc}. For each of examples, we provide the traces of three random experiments. For all examples, the Gibbs sampling applied to the $\mat{A}\mat{B}^T$ model do not have satisfactory mixing property---after several thousands iterations, the traces of Markov chains clearly suggest that the chains are still not in the stationary phase. In comparison, all the geodesic HMC counterparts can produce well mixed chains after several hundreds iterations. Although each iteration of the geodesic HMC is computationally more expensive than that of the Gibbs, the geodesic HMC is able to significantly improve the convergence of the MCMC sampling in these examples.

%% file: main.bbl
\begin{thebibliography}{10}

\bibitem{ahn2015large}
Sungjin Ahn, Anoop Korattikara, Nathan Liu, Suju Rajan, and Max Welling.
\newblock Large-scale distributed bayesian matrix factorization using
  stochastic gradient mcmc.
\newblock In {\em Proceedings of the 21th ACM SIGKDD international conference
  on knowledge discovery and data mining}, 2015.

\bibitem{arnold2013}
Vladimir~Igorevich Arnol'd.
\newblock {\em Mathematical methods of classical mechanics}, volume~60.
\newblock Springer Science \& Business Media, 2013.

\bibitem{bennett2007netflix}
James Bennett, Stan Lanning, et~al.
\newblock The netflix prize.
\newblock In {\em Proceedings of KDD cup and workshop}, 2007.

\bibitem{byrne2013}
Simon Byrne and Mark Girolami.
\newblock Geodesic monte carlo on embedded manifolds.
\newblock {\em Scandinavian Journal of Statistics}, 40(4):825--845, 2013.

\bibitem{candes2010power}
Emmanuel~J Cand{\`e}s and Terence Tao.
\newblock The power of convex relaxation: Near-optimal matrix completion.
\newblock {\em IEEE Transactions on Information Theory}, 56(5):2053--2080,
  2010.

\bibitem{chen2014stochastic}
Tianqi Chen, Emily Fox, and Carlos Guestrin.
\newblock Stochastic gradient hamiltonian monte carlo.
\newblock In {\em International conference on machine learning}, 2014.

\bibitem{DeGorodetsky2020}
Saibal De, Hadi Salehi, and Alex Gorodetsky.
\newblock Efficient mcmc sampling for bayesian matrix factorization by breaking
  posterior symmetries.
\newblock {\em arXiv preprint arXiv:2006.04295}, 2020.

\bibitem{Diaconis2013}
Persi Diaconis, Susan Holmes, and Mehrdad Shahshahani.
\newblock Sampling from a manifold.
\newblock In {\em Advances in modern statistical theory and applications: a
  Festschrift in honor of Morris L. Eaton}, pages 102--125. Institute of
  Mathematical Statistics, 2013.

\bibitem{Dua:2019}
Dheeru Dua and Casey Graff.
\newblock {UCI} machine learning repository, 2017.

\bibitem{duane1987hybrid}
Simon Duane, Anthony~D Kennedy, Brian~J Pendleton, and Duncan Roweth.
\newblock Hybrid monte carlo.
\newblock {\em Physics letters B}, 195(2):216--222, 1987.

\bibitem{edelman1998}
Alan Edelman, Tom{\'a}s~A Arias, and Steven~T Smith.
\newblock The geometry of algorithms with orthogonality constraints.
\newblock {\em SIAM journal on Matrix Analysis and Applications},
  20(2):303--353, 1998.

\bibitem{federer2014geometric}
Herbert Federer.
\newblock {\em Geometric measure theory}.
\newblock Springer, 2014.

\bibitem{MCMC:GiCal_2011}
M.~Girolami and B.~Calderhead.
\newblock Riemann manifold {L}angevin and {H}amiltonian {M}onte {C}arlo
  methods.
\newblock {\em Journal of the Royal Statistical Society: Series B (Statistical
  Methodology)}, 73(2):123--214, 2011.

\bibitem{harper2015movielens}
F~Maxwell Harper and Joseph~A Konstan.
\newblock The movielens datasets: History and context.
\newblock {\em ACM transactions on interactive intelligent systems (TIIS)},
  5(4):1--19, 2015.

\bibitem{hastie2015matrix}
Trevor Hastie, Rahul Mazumder, Jason~D Lee, and Reza Zadeh.
\newblock Matrix completion and low-rank svd via fast alternating least
  squares.
\newblock {\em The Journal of Machine Learning Research}, 16(1):3367--3402,
  2015.

\bibitem{higuera2015self}
Clara Higuera, Katheleen~J Gardiner, and Krzysztof~J Cios.
\newblock Self-organizing feature maps identify proteins critical to learning
  in a mouse model of down syndrome.
\newblock {\em PloS one}, 10, 2015.

\bibitem{hoffman2014no}
Matthew~D Hoffman, Andrew Gelman, et~al.
\newblock The no-u-turn sampler: adaptively setting path lengths in hamiltonian
  monte carlo.
\newblock {\em J. Mach. Learn. Res.}, 15(1):1593--1623, 2014.

\bibitem{lim2007variational}
Yew~Jin Lim and Yee~Whye Teh.
\newblock Variational bayesian approach to movie rating prediction.
\newblock In {\em Proceedings of KDD cup and workshop}, 2007.

\bibitem{liu2001monte}
Jun~S Liu.
\newblock {\em Monte Carlo strategies in scientific computing}, volume~10.
\newblock Springer, 2001.

\bibitem{mazumder2010spectral}
Rahul Mazumder, Trevor Hastie, and Robert Tibshirani.
\newblock Spectral regularization algorithms for learning large incomplete
  matrices.
\newblock {\em The Journal of Machine Learning Research}, 11:2287--2322, 2010.

\bibitem{nakajima2011theoretical}
Shinichi Nakajima and Masashi Sugiyama.
\newblock Theoretical analysis of bayesian matrix factorization.
\newblock {\em Journal of Machine Learning Research}, 12, 2011.

\bibitem{nakajima2013global}
Shinichi Nakajima, Masashi Sugiyama, S.~Derin Babacan, and Ryota Tomioka.
\newblock Global analytic solution of fully-observed variational bayesian
  matrix factorization.
\newblock {\em Journal of Machine Learning Research}, 14, 2013.

\bibitem{neal2011mcmc}
Radford~M Neal et~al.
\newblock Mcmc using hamiltonian dynamics.
\newblock {\em Handbook of markov chain monte carlo}, 2(11):2, 2011.

\bibitem{rai2014scalable}
Piyush Rai, Yingjian Wang, Shengbo Guo, Gary Chen, David Dunson, and Lawrence
  Carin.
\newblock Scalable bayesian low-rank decomposition of incomplete multiway
  tensors.
\newblock In {\em International Conference on Machine Learning}, 2014.

\bibitem{raiko2007principal}
Tapani Raiko, Alexander Ilin, and Juha Karhunen.
\newblock Principal component analysis for large scale problems with lots of
  missing values.
\newblock In {\em European Conference on Machine Learning}, 2007.

\bibitem{robert1999monte}
Christian~P Robert and George Casella.
\newblock {\em Monte Carlo statistical methods}, volume~2.
\newblock Springer, 1999.

\bibitem{salakhutdinov2008bayesian}
Ruslan Salakhutdinov and Andriy Mnih.
\newblock Bayesian probabilistic matrix factorization using markov chain monte
  carlo.
\newblock In {\em Proceedings of the 25th international conference on Machine
  learning}, 2008.

\bibitem{zhao2015bayesian1}
Qibin Zhao, Liqing Zhang, and Andrzej Cichocki.
\newblock Bayesian cp factorization of incomplete tensors with automatic rank
  determination.
\newblock {\em IEEE transactions on pattern analysis and machine intelligence},
  37, 2015.

\bibitem{zhao2015bayesian2}
Qibin Zhao, Guoxu Zhou, Liqing Zhang, Andrzej Cichocki, and Shun-Ichi Amari.
\newblock Bayesian robust tensor factorization for incomplete multiway data.
\newblock {\em IEEE transactions on neural networks and learning systems}, 27,
  2015.

\end{thebibliography}
